\documentclass[10pt,twocolumn,letterpaper]{article}

\usepackage[pagenumbers]{cvpr} 

\usepackage{graphicx}
\usepackage[permil]{overpic}
\usepackage{makecell}
\usepackage{multirow}
\usepackage{subcaption}
\usepackage{wrapfig}

\usepackage[dvipsnames]{xcolor}

\newcommand{\segtoreg}{\textsf{Seg2Reg}\xspace}
\newcommand{\codebase}{\textsf{PanoLayoutStudio}\xspace}

\newcommand{\Real}{\mathbb{R}}
\newcommand{\img}{\mathcal{I}}
\newcommand{\eqpt}{\mathbf{u}}
\newcommand{\polyvert}{\mathbf{v}}
\newcommand{\polyvertprimary}{\polyvert^{\mathrm{(primary)}}}

\newcommand{\polyvertceiling}{\polyvert^{\mathrm{(ceiling)}}}
\newcommand{\height}{h}
\newcommand{\encoder}{\mathrm{Enc}}
\newcommand{\transform}{\mathrm{T}}
\newcommand{\feature}{\mathcal{F}}
\newcommand{\featureup}{\hat{\feature}}
\newcommand{\densitymap}{\mathcal{D}}
\newcommand{\densitylogitmap}{\widetilde{\densitymap}}
\newcommand{\density}{\rho}
\newcommand{\densitylogit}{\tilde{\density}}
\newcommand{\zfloor}{z^{\mathrm{(floor)}}}
\newcommand{\zceiling}{z^{\mathrm{(ceiling)}}}
\newcommand{\ray}{\mathbf{r}}
\newcommand{\rayo}{\ray_\mathrm{o}}
\newcommand{\rayd}{\ray_\mathrm{d}}
\newcommand{\dist}{d}
\newcommand{\npts}{K}
\newcommand{\seglen}{\Delta}
\newcommand{\interp}{\mathrm{Interp}}
\newcommand{\eqproj}{\mathrm{EqProj}}
\newcommand{\softplus}{\mathrm{Softplus}}
\newcommand{\sigmoid}{\mathrm{Sigmoid}}
\newcommand{\rendpoly}{\mathrm{RendPoly}}
\newcommand{\hit}{\mathrm{Hit}}
\newcommand{\nsecondary}{N^{\mathrm{(secondary)}}}

\newcommand{\loss}{\mathcal{L}}
\newcommand{\Lprimary}{\loss^\mathrm{(pri.)}}
\newcommand{\Lsecondary}{\loss^\mathrm{(2nd)}}
\newcommand{\Lseg}{\loss^\mathrm{(seg.)}}

\newcommand{\degree}{$^{\circ}\,$}

\newcommand{\ourimpl}{\textcolor[HTML]{D8B260}{\textsuperscript{\tiny $\blacklozenge$}}}

\newcommand{\gold}[1]{\colorbox[rgb]{1.0, 0.874, 0}{#1}}

\definecolor{cvprblue}{rgb}{0.21,0.49,0.74}
\usepackage[pagebackref,breaklinks,colorlinks,citecolor=cvprblue]{hyperref}

\def\paperID{9800} 
\def\confName{CVPR}
\def\confYear{2024}

\title{Seg2Reg: Differentiable 2D Segmentation to 1D Regression Rendering for 360 Room Layout Reconstruction}

\author{
Cheng Sun$^{1}$ \qquad
Wei-En Tai$^{2}$ \qquad
Yu-Lin Shih$^{2}$ \qquad
Kuan-Wei Chen$^{2}$ \\
Yong-Jing Syu$^{2}$ \qquad
Kent Selwyn The$^{2}$ \qquad
Yu-Chiang Frank Wang$^{1,3}$ \qquad
Hwann-Tzong Chen$^{2,4}$ \\
{\small $^{1}$NVIDIA} \quad
{\small $^{2}$National Tsing Hua University} \quad
{\small $^{3}$National Taiwan University} \quad
{\small $^{4}$Aeolus Robotics}
}

\begin{document}
\maketitle
\begin{abstract}
State-of-the-art single-view 360\degree room layout reconstruction methods formulate the problem as a high-level 1D (per-column) regression task. On the other hand, traditional low-level 2D layout segmentation is simpler to learn and can represent occluded regions, but it requires complex post-processing for the targeting layout polygon and sacrifices accuracy. We present \segtoreg to render 1D layout depth regression from the 2D segmentation map in a differentiable and occlusion-aware way, marrying the merits of both sides. Specifically, our model predicts floor-plan density for the input equirectangular 360\degree image. Formulating the 2D layout representation as a density field enables us to employ `flattened' volume rendering to form 1D layout depth regression. In addition, we propose a novel 3D warping augmentation on layout to improve generalization. Finally, we re-implement recent room layout reconstruction methods into our codebase for benchmarking and explore modern backbones and training techniques to serve as the strong baseline. Our model significantly outperforms previous arts. The code will be made available upon publication.
\end{abstract}
\vspace{-2em}
\section{Introduction}
\label{sec:intro}

Room layout estimation is one of the fundamental vision problems toward scene understanding.
The goal is to reconstruct the outermost room structure, usually comprising the floor, ceiling, walls, and sometimes columns and beams.
Room layout is crucial in various indoor tasks, such as holistic 3D reconstruction~\cite{ZhangSTX14,LiYSWSLYZGSBYXS21,ZhangCCLZB021,YehLHZXHSC22}, image synthesis~\cite{XuZXTG21,GaoCSC22}, floor-plan estimation~\cite{CruzHLKBK21,SolarteLWTS22}, and extreme baseline SfM~\cite{ShabaniSOFF21,HutchcroftLBWWK22}.
Automatic layout annotation for panoramas is also a sought-after feature in real estate portals.

\begin{figure}
    \centering
    \includegraphics[trim=0 550 30 0, clip,width=\linewidth]{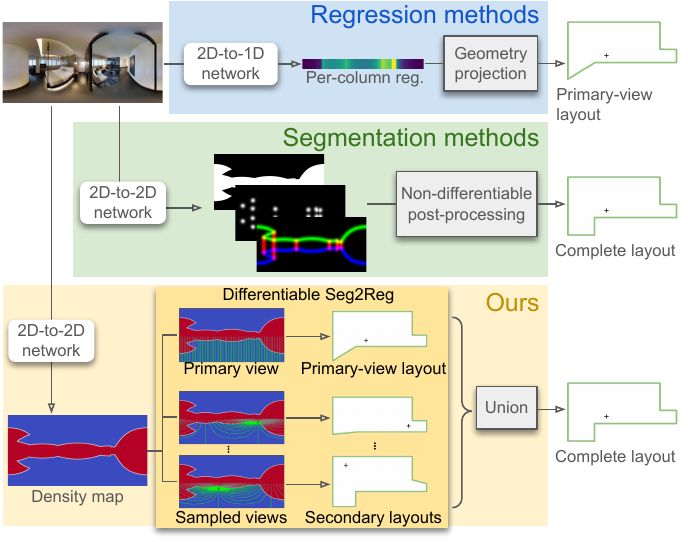}
    \caption{
        \textcolor[HTML]{2854C5}{\bf Regression methods} directly predict layout geometry.
        A powerful 2D-to-1D decoder is essential to capture high-level cues within an image column.
        The regressed layout pertains solely to the visible region from the camera origin (dubbed primary-view layout).
        \textcolor[HTML]{48752C}{\bf Segmentation methods} predict lower-level per-pixel probabilities of layout facades, corners, and boundaries. While capable of modeling occluded regions, they require a non-differentiable post-heuristic to convert segmentation to layout geometry.
        Our proposed \textcolor[HTML]{B89230}{\bf \segtoreg} aims to synergize the strengths of both.
        We re-formulate 2D layout representation as 2D density field via projecting pixels onto the floor or ceiling.
        We use the classical volume rendering technique to render depth on the density map, which is differentiable and occlusion-aware.
        The 1D depth maps rendered from the primary view and the sampled secondary views directly outline the layout polygons.
    }
    \label{fig:teaser}
\end{figure}

Traditional deep-learning methods view room-layout estimation as semantic segmentation tasks for perspective images~\cite{ZhaoLYGCZ17,IzadiniaSS17,LeeBMR17} or panoramas~\cite{ZouCSH18,YangWPWSC19,PintoreAG20,ZouSPCSWCH21}.
The downside of previous segmentation-based methods is that they need heuristic post-processing steps, which introduce errors and gaps between the training objective and the targeted layout outcomes.
Recent advancements in 360\degree room-layout estimation involve training deep models to regress the \emph{boundary}~\cite{SunHSC19,SunSC21} or \emph{distance to layout wall}~\cite{WangYSCT21,JiangXXZ22} for each image column (\ie, 1D regression).
The regressed geometry information can then be directly projected onto the floor to form a layout floor-plan polygon.
Despite achieving state-of-the-art accuracy, regression-based methods require a robust and large decoder to learn to capture global-scale information. Moreover, these methods still rely on post-processing heuristics to infer occluded regions.

Our method, \segtoreg, enables the differentiable rendering of 1D layout depth (\ie, distance to wall) to form floor-plan polygon from 2D segmentation-based representation on a 360\degree image (\cref{fig:teaser}).
The key insight is to re-formulate the 2D layout representation as 2D floor-plan density field, allowing us to employ the classical volume rendering technique~\cite{KajiyaH84,Max95a}, which has recently gained great success in 3D NeRF~\cite{MildenhallSTBRN20} modeling, to compute geometry information in a soft and differentiable manner.
Like NeRF, our method involves volume rendering of rays, but in our case, the `ray' trajectory is on the predicted 2D density logit map rather than in 3D space, and the ray should be bent as per the 360\degree imaging.

Notably, we show that it is crucial to train our segmentation-based representation with the regression objective, which aligns more directly with our intended task goals.
Solely training with the segmentation objective results in lower accuracy, suggesting that our \segtoreg is the missing piece in earlier segmentation-based methods~\cite{ZouCSH18,YangWPWSC19,PintoreAG20,ZouSPCSWCH21} to achieve state-of-the-art performance. 
Beyond the visible layout, the volume rendering algorithm and our predicted density map are occlusion-aware, so we can also directly render the floor-plan polygon vertices for the occluded region.

Our codebase, dubbed \codebase, implements various modern backbones and training techniques.
Additionally, we extend the widely-used PanoStretch~\cite{SunHSC19} data augmentation to allow for more flexible random adjustments on layout corners.
Finally, we reproduce most of the recent methods in our codebase, establishing a stronger baseline for fair comparisons and providing a resource for future works to reuse or recombine various components.

In summary, our contributions are as follows:
{\textit{i)}} The proposed \segtoreg differentiably renders 1D regression from 2D segmentation, marrying the merits of both formulations and resulting in a smaller yet stronger model.
{\textit{ii)}} Our method directly estimates occluded layouts without relying on heuristic post-processing.
{\textit{iii)}} We introduce a new data-augmentation scheme that allows for flexible layout 3D adjustments. 
{\textit{iv)}} We make a system-level contribution by modernizing the backbones and the training recipes for layout estimation and reproducing previous methods in our codebase---\codebase, which boosts all methods and eases future efforts with reusable modules.

\section{Related work}
\label{sec:relatedwork}
\paragraph{Panorama layout estimation.}
Early layout estimation takes perspective images as input~\cite{ZhaoLYGCZ17,IzadiniaSS17,LeeBMR17}, while recent approaches increasingly focus on panoramic images and often rely on predicting distinct scene structures, such as ceiling, floor boundaries, or wall corners~\cite{ZhangSTX14, ZouCSH18, SunHSC19, SunSC21, ZouSPCSWCH21, Fernandez-Labrador20}. LayoutNet~\cite{ZouCSH18} is the first to predict boundaries and corners on a single panoramic image using deep neural networks.

HorizonNet~\cite{SunHSC19} is the pioneering method to reformulate this task as a per-column regression problem, utilizing a powerful deep model to regress boundary positions instead of the conventional low-level heatmap layout encoding.
Succeeding enhancements are made from both the model architecture~\cite{SunSC21} and the layout representation~\cite{WangYSCT21}.
LGTNet~\cite{JiangXXZ22} takes a step further by employing a transformer-based architecture with an improved layout formulation, which consists of layout depth and layout height, ultimately achieving state-of-the-art quality.


Since LayoutNet~\cite{ZouCSH18}, segmentation-based works also seek to boost quality by improving layout encoding~\cite{YangWPWSC19} and network architecture~\cite{PintoreAG20,ZouSPCSWCH21}.
In contrast to regression-based methods, where model predictions can be directly projected into 3D, segmentation-based methods heavily depend on post-heuristic to convert heatmap predictions into layout geometry.
We find that the disconnection between the training objective and the final outcome is a bottleneck in achieving superior results, and we propose to reformulate the probability heatmap as a 2D density field so that we can employ differentiable rendering for the geometric regression properties as well.

\paragraph{Neural radiance field.}
NeRF~\cite{MildenhallSTBRN20} is the de facto method for multiview 3D reconstruction research in recent years.
It combines MLP and volume rendering~\cite{KajiyaH84,Max95a} to model the density field and color field of a scene.
Subsequent works~\cite{SunSC22,Fridovich-KeilY22,MullerESK22,ChenXGYS22} show that MLP is not necessary while some grid-based representations can also work well.
Inspired by their success, we train an ultra-light model to predict a low-level 2D density field and leverage volume rendering to accumulate density into high-level geometric properties, forming 2D polygons directly and differentiably.


\paragraph{Data augmentation.}
The progress of data augmentations for perspective images~\cite{ZhangCDL18, YunHCOYC19, DaboueiSTN21, CubukZMVL19, LimKKKK19, CubukZSL20, HofferBHGHS20, Zhong0KL020} is rapid in recent years as it plays a crucial role in achieving better results.
Among these augmentations, geometric-based data augmentation is found to be especially beneficial~\cite{CubukZSL20}.
Unfortunately, geometric-based data augmentation for 360\degree layout estimation is rather limited.
PanoStretch~\cite{SunHSC19} randomly adjusts layout aspect ratio.
PanoMixSwap~\cite{abs-2309-09514} uses a generative model to mix furniture, backgrounds, and layout structures from different 360\degree images, which is, however, time-consuming.
The challenge is that existing geometric augmentation, such as image y-translation, breaks the underlying ground-truth layout sanity, which makes them inapplicable.
We present a principled solution to perform geometric data augmentation for 360\degree layout, enabling the generation of a more diverse data distribution beyond existing techniques.


\section{Approach}
\label{sec:approach}

The input is a 360\degree panoramic image $\img\in\Real^{H\times W\times 3}$ under equirectangular projection.
The target room layout can be represented by a sequence of 2D coordinates $\{\polyvert_i^*\}_{i=1}^{K}$, which forms a $K$-edge polygon outlining the floor plan, with a scalar $\height^*$ for the layout height.
To fix the scale ambiguity, we follow the literature to rescale the camera-to-floor distance to $1.6$ meters.
In \cref{ssec:segtoreg}, we introduce our novel layout representation and floor-plan polygon rendering.
\cref{ssec:arch} details our model design.
\cref{ssec:newaug} presents a new principle to perform geometric augmentation for 360\degree layout tasks.
Finally, \cref{ssec:codebase} presents our codebase.


\subsection{Seg2Reg}
\label{ssec:segtoreg}
\begin{figure}
    \centering
    \includegraphics[trim=0 295 330 0, clip,width=\linewidth]{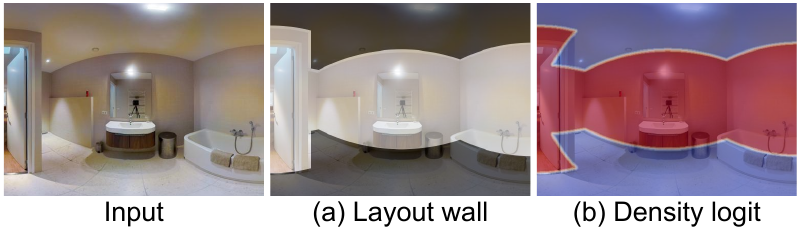}
    \caption{
        {\bf Layout wall vs. our density logit.}
        (a) Layout wall segmentation delineates the room layout in the image space.
        (b) We train the model to predict a pixel density map when projected to the floor or ceiling planes.
        The density map enables us to render depth in a soft and differentiable way (\cref{ssec:segtoreg}).
    }
    \label{fig:dens_vs_wallseg}
\end{figure}

Our layout representation is a pixel-level density logit map $\densitylogitmap \in \Real^{H\times W\times 1}$.
We can use $\softplus$ to convert the density logit into non-negative density:
\begin{equation} \label{eq:softplus}
    \densitymap = \log\left(1 + \exp\left(\densitylogitmap\right)\right) ~.
\end{equation}
When projected to the floor or ceiling, the density $\densitymap_q$ indicates a pixel $q$ is exterior (\ie, high density) or interior (\ie, low density) to the room layout.
Unlike segmenting layout walls, the density map needs to `see through' walls that might be blocking the view for an interior region (\cref{fig:dens_vs_wallseg}).

\paragraph{Overview.}
We set the floor plane position at $\zfloor{=}1.6$ and a temporary ceiling plane position at $\zceiling{=}{-}1$ (we use z-down positive world coordinate system).
The upper- and bottom-half of $\densitymap$ (\ie, first and last $\frac{H}{2}$ image rows) estimate the density on the ceiling and floor planes, respectively.
In the following, we introduce our algorithm to render the 2D layout polygon on the floor plane.
The same algorithm can be applied to render the ceiling-projected polygon.
Following standard practice~\cite{SunHSC19,SunSC21,WangYSCT21}, we take the polygon on the floor as the main 2D layout outline, while the ceiling polygon is only used to infer the layout height $\height$.

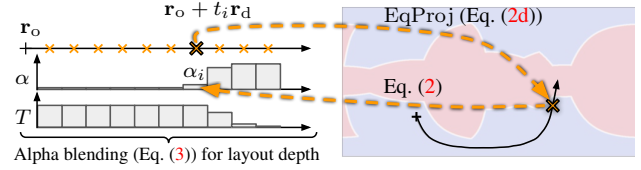
\begin{figure}
    \centering
    \begin{overpic}[trim=0 320 410 0, clip,width=\linewidth, clip]{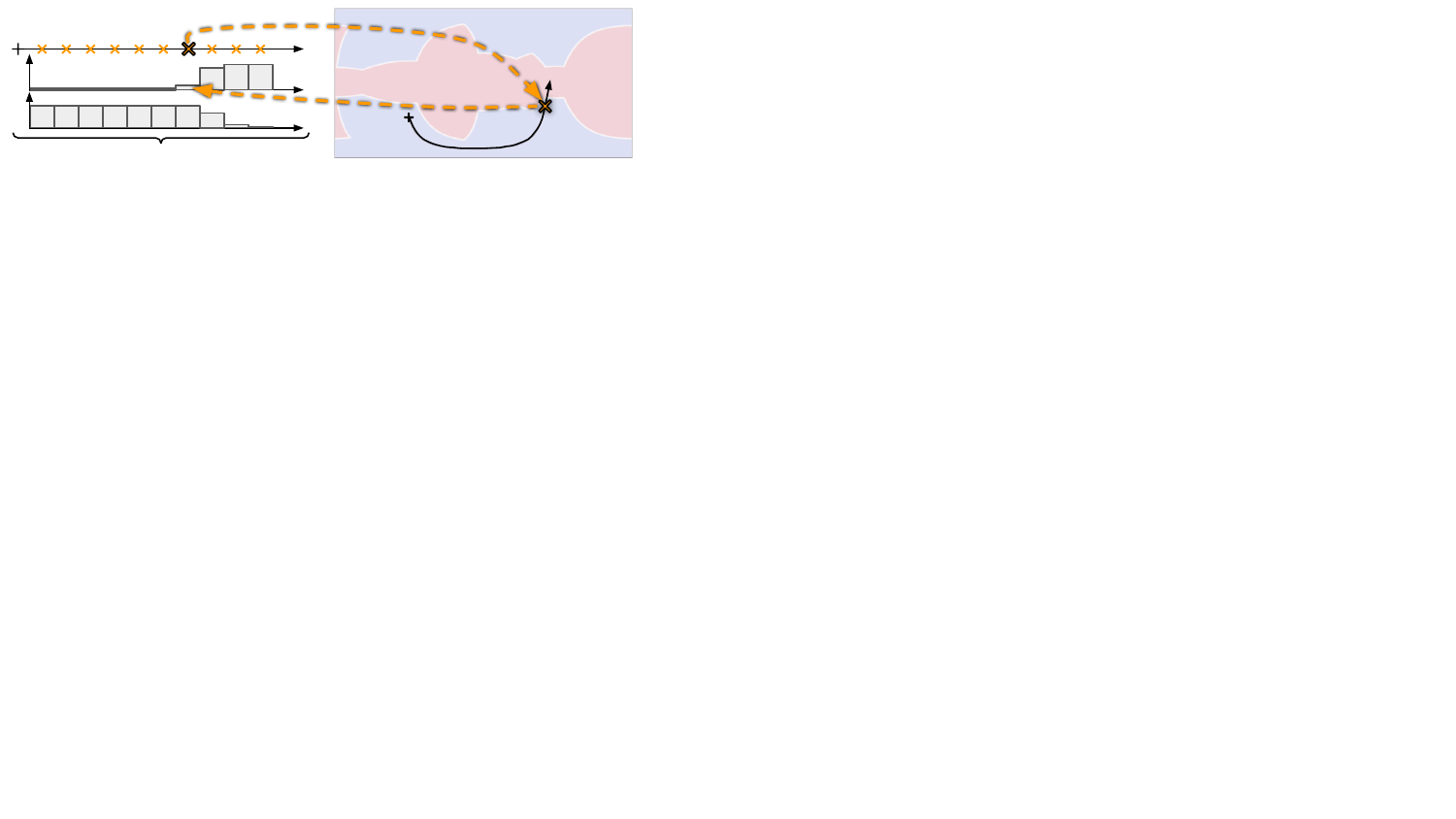}
    \put(20,220){\footnotesize $\rayo$}
    \put(250,250){\footnotesize $\rayo + t_i \rayd$}

    \put(600,238){\footnotesize $\eqproj$ (\cref{subeq:eqproj})}
    \put(600,125){\footnotesize \cref{eq:alpha}}
    \put(280,150){\footnotesize $\alpha_i$}

    \put(12,140){\footnotesize $\alpha$}
    \put(12,75){\footnotesize $T$}

    \put(10,18){\scriptsize Alpha blending (\cref{eq:alpha_blending}) for layout depth}
    \end{overpic}
    \caption{
    {\bf Visualization of the `flattened' volume rendering from a given ray.}
    Please refer to \cref{ssec:segtoreg} for the details.
    }
    \label{fig:volrend}
\end{figure}

\paragraph{`Flattened' volume rendering on the 2D floor plan.}
We illustrate the rendering of a ray in \cref{fig:volrend}.
Given a camera position $\rayo$ and a unit vector of the ray direction $\rayd$ on the 2D layout of the floor plan, we want to render the expected distance $\dist$ to the layout exterior based on the estimated $\densitylogitmap$.
We first sample a series of $K$ points on the ray, denoted by $\{\rayo + t_i \rayd\}_{i=1}^\npts$, ordered from nearest to farthest.
The opacity $\alpha_i \in [0, 1]$ of the $i$-th sampled point is
\begin{subequations} \label{eq:alpha}
\begin{align}
    \alpha_i &= 1 - \exp\left(-\density_i \seglen_i \right) ~, \label{subeq:ptalpha}\\
    \density_i &= \softplus(\densitylogit_i) ~, \label{subeq:ptsoftplus}\\
    \densitylogit_i &= \interp\left( \eqpt_i, \densitylogitmap \right) ~, \\
    \eqpt_i &= \eqproj\left(\left[\rayo + t_i \rayd, \zfloor\right]\right) ~, \label{subeq:eqproj}
\end{align}
\end{subequations}
where $\eqproj(\cdot)$ projects the 3D point to equirectangular image coordinate $\eqpt$, $\interp(\eqpt, \densitylogitmap)$ bilinearly interpolates the density $\density$ of point $\eqpt$ on $\densitylogitmap$, and $\seglen_i$ is the spherical distance of the $i$-th ray segment projected to a unit sphere.
Please refer to the supplementary for the details of the coordinate system and transformation in this work.
We apply the $\softplus$ activation after the interpolation (post-activation) for sharper decision boundary~\cite{SunSC22}.
Finally, the distance to layout boundary $d$ is computed by alpha blending:
\begin{equation} \label{eq:alpha_blending}
    d = \sum\nolimits_{i=1}^\npts T_i \alpha_i t_i ~, \text{where}~ T_i = \prod\nolimits_{j=1}^{i-1} \left(1 - \alpha_j\right) ~.
\end{equation}
The differentiable term $(T_i \alpha_i)$ is the probability of the ray stopping at point $(\rayo + t_i \rayd)$.
Hereinafter, we use
\begin{equation}
    \hit(\rayo, \rayd){=}\rayo + d \rayd
\end{equation}
to denote the expected ray-polygon intersection by rendering depth.
The dependent density logit map $\densitylogitmap$ and the plane position $z$ of the function are omitted for brevity.

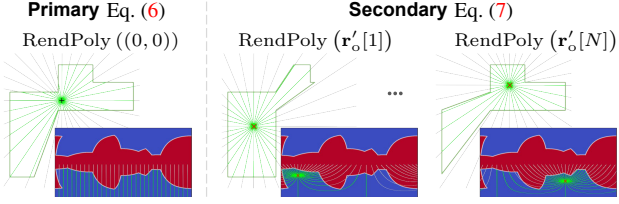
\begin{figure}
    \centering
    \begin{overpic}[trim=0 280 330 0, clip,width=\linewidth, clip]{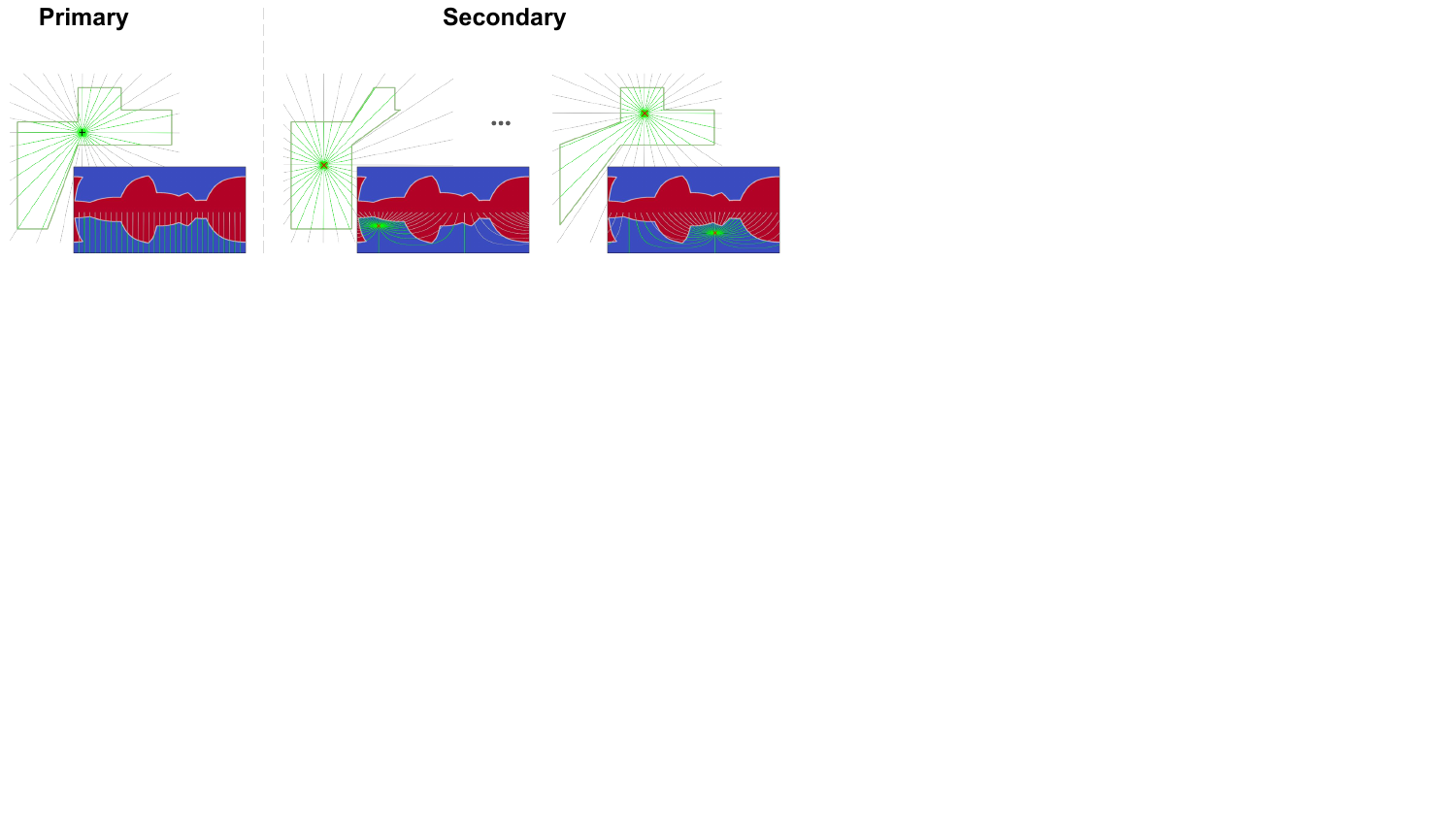}
    \put(175,287){\footnotesize \cref{eq:poly_primary}}
    \put(730,287){\footnotesize \cref{eq:poly_2nd}}

    \put(40,240){\scriptsize $\rendpoly\left( (0,0) \right)$}
    \put(385,240){\scriptsize $\rendpoly\left( \rayo'[1] \right)$}
    \put(730,240){\scriptsize $\rendpoly\left( \rayo'[N] \right)$}
    \end{overpic}
    \caption{
    {\bf Primary and secondary layout polygons rendering.}
    Our model predicts the density logit (\cref{eq:softplus}).
    Given a camera position $\rayo$ on the floor plan, we employ `flattened' volume rendering (\cref{eq:alpha,eq:alpha_blending}) to render a layout polygon, $\rendpoly\left( \rayo \right)$, based on the predicted density logit map.
    }
    \label{fig:rendpoly}
\end{figure}

\paragraph{Primary layout polygon.}
With the flattened volume rendering algorithm, we can now directly render an $M$-edge polygon from a given camera center $\rayo$:
\begin{equation}
    \rendpoly_M\left(\rayo\right) = 
    \left\{
    \hit\left(\rayo,~ \rayd^{(M)}[i]\right)
    \right\}_{i=1}^M ~,
\end{equation}
where $\rayd^{(M)}$ is a set of $M$ unit-vectors uniformly spacing around a circle.
We synthesize a $W$-edge primary polygon by placing a camera at $(0, 0)$:
\begin{equation} \label{eq:poly_primary}
    \left\{\polyvertprimary_i\right\}_{i=1}^W = \rendpoly_W\left((0,0)\right) ~,
\end{equation}
so the distance to each of the $W$ vertices corresponds to the layout depth of the source image column.
In case of no self-occlusion, the rendered primary layout polygon is capable of representing the whole room. 
We illustrate the layout polygon rendering in \cref{fig:rendpoly}.

\paragraph{Secondary layout polygons.}
To inference the occlusion region, we sample additional cameras $\rayo'$ from the interior region to render a set of $\nsecondary$ layout polygons
\begin{equation} \label{eq:poly_2nd}
    \{\rendpoly_W\left( \rayo'[i] \right)\}_{i=1}^{\nsecondary} ~.
\end{equation}
The interior region is determined from the primary layout polygon during testing, while we sample from the ground-truth interior region during training.
We can compute the union over polygons to merge the secondary polygons into the primary ones.
We also implement a rendering noise (due to numerical integration) robust algorithm based on minimum spanning tree and tree diameter, which is detailed in the supplementary material.

\paragraph{Layout height inference.}
The value of $\zfloor$ is fixed, so we only have to estimate the ceiling plane's $z$ position.
First, the ceiling polygon $\{\polyvertceiling\}_{i=1}^W$ is rendered in the same way as the floor in \cref{eq:poly_primary} but on the temporary ceiling plane position $\zceiling$ instead.
We then find the scale
\begin{equation}
    s^* = \min_s \sum_{i=1}^W \left(\left\| \polyvertprimary_i\right\| - s \left\|\polyvertceiling_i \right\|\right)^2
\end{equation}
that aligns the ceiling polygon to the primary floor polygon.
The formula of the layout height by solving least-squares is
\begin{equation}
    h = \zfloor - \zceiling \frac{\sum_{i=1}^W \left\|\polyvertprimary_i\right\|\left\|\polyvertceiling_i\right\|}{\sum_{i=1}^W \left\|\polyvertceiling_i\right\|^2} ~.
\end{equation}

\paragraph{Relation to binary segmentation.}
We can merge the \cref{subeq:ptalpha,subeq:ptsoftplus} as
\begin{equation}
\begin{split}
    \alpha &= 1 - \exp\left(-\softplus(\densitylogit) \seglen \right) \\
    &= 1 - \exp\left(-\log\left(1 + \exp\left(\densitylogit\right)\right)\seglen\right) \\
    &= 1 - \left(1 + \exp\left(\densitylogit\right)\right)^{-\seglen} \\
    &= 1 - \sigmoid\left(-\densitylogit\right)^\seglen ~,
\end{split}
\end{equation}
where the subscripts are omitted for brevity.
The spherical distance of a pixel height on the equirectangular image is $\frac{\pi}{H}$.
We re-scale $\seglen$ by $\frac{H}{\pi}$ so the opacity of a vertical ray segment centered at a pixel $q$ can be simplified to
\begin{equation}
    \alpha_q = 1- \sigmoid\left( -\densitylogitmap_q \right) = \sigmoid\left( \densitylogitmap_q \right) ~.
\end{equation}
We can see that the predicted density logit map $\densitylogitmap$ can be reduced to the \emph{binary} segmentation logit if we do not apply rendering, which enables us to apply segmentation loss as a training auxiliary.

\paragraph{Training objective.}
For each ray to render the primary and secondary layout polygons, we also compute their depth to the ground-truth polygon.
Minimizing the difference between the rendered and ground-truth depth directly introduces ambiguity, as there exists an infinite number of weight distributions in the alpha blending (\cref{eq:alpha_blending}) that can yield the same result.
Instead, we derive a compact weight distribution $w^*$ that renders the ground-truth depth with only the two nearest points having weight (detailed in the supplementary).
We directly guide the alpha blending weight distribution of a ray in \cref{eq:alpha_blending} via cross-entropy loss:
\begin{equation}
     - w^*_{K+1}\log\left(T_{K+1}\right) - \sum\nolimits_{i=1}^K w^*_i \log\left(T_i\alpha_i\right) ~.
\end{equation}
We apply the cross-entropy loss to the rendered primary and secondary layouts, and the losses are denoted as $\Lprimary$ and $\Lsecondary$, respectively.
The cross-entropy loss for the alpha blending weight mainly focuses on the interior and boundary regions.
To prevent random results in the far exterior region, we also apply binary segmentation loss $\Lseg$ to the predicted density logit.
Our final training objective is
\begin{equation}
    \loss = w_1 \, \Lprimary + w_2 \, \Lsecondary + w_3 \, \Lseg ~.
\end{equation}


\subsection{Network architecture}
\label{ssec:arch}

We first detail our network architecture for predicting the density logit map and then compare our approach with closely related work.

\paragraph{Backbone.}
The backbone predicts a feature pyramid in four levels for the input image:
\begin{equation}
    \{\feature_i \in \Real^{H_i\times W_i\times C_i}\}_{i=1}^4 = \encoder(\img) ~,
\end{equation}
where $H_i{=}\frac{H}{2^{i+1}}$, $W_i{=}\frac{W}{2^{i+1}}$, and $C_i$ is the backbone's channel dimension.

\paragraph{Segmentation-based 2D decoder.}
We adopt an all-MLP decoder design~\cite{XieWYAAL21} to predict density logit map:
\begin{subequations}
\label{eq:decoder2d}
\begin{align}
    \featureup_i &= \mathrm{Upsample}_{(H,W)}\left(\mathrm{Linear}_{C_i\rightarrow D}(\feature_i)\right) ~, \\
    \densitylogitmap &= \mathrm{Linear}_{D\rightarrow 1}\left(\mathrm{GELU}\left( \sum\nolimits_{i=1}^4 \featureup_i \right) \right) ~,
\end{align}
\end{subequations}
where $\mathrm{Linear}_{C_i\rightarrow C_o}(\cdot)$ is a linear layer mapping the number of latent channels from $C_i$ to $C_o$, and $\mathrm{Upsample}_{(H,W)}(\cdot)$ bilinearly interpolates the spatial size to $(H,W)$.


\paragraph{Discussions about top-down view models.}
Our \segtoreg can also be applied to top-down view (\ie, ceiling view, floor-plan view, or bird's-eye view) segmentation-based models~\cite{YangWPWSC19,PintoreAG20}.
However, we find it hard to choose an appropriate perspective field-of-view as small FoVs miss farther walls while large FoVs limit the space for closer regions.
We mainly follow recent state-of-the-art to use equirectangular view and leave our method's application to perspective view for potential future explorations.


\subsection{Layout 3D warping}
\label{ssec:newaug}
\begin{figure}
    \centering
    \begin{overpic}[trim=0 273 240 0, clip, width=\linewidth]{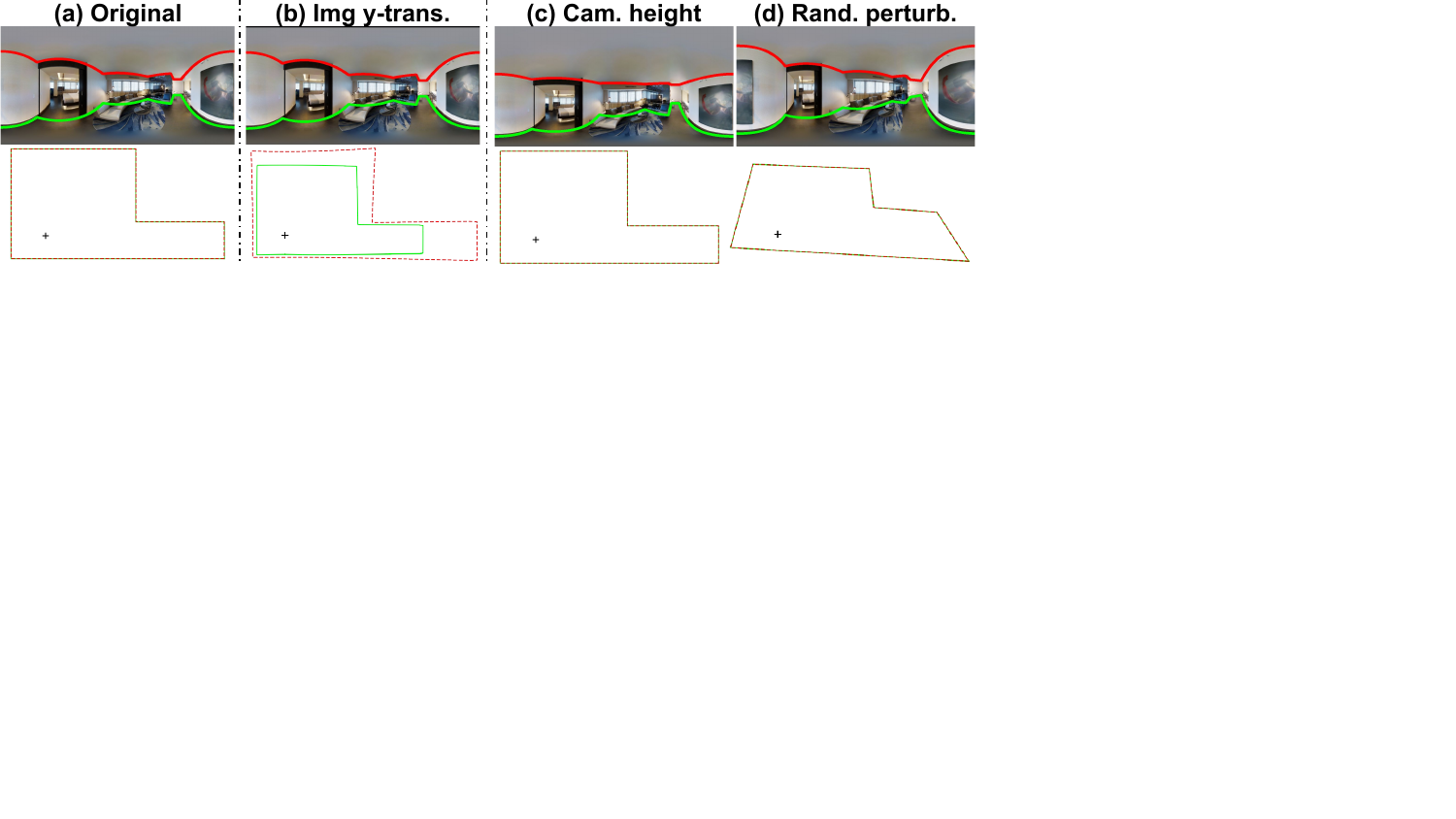}
    \end{overpic}
    \caption{
    {\bf Visualization of layout warping.}
    (b) Some commonly used geometric data augmentations, like image y-translation, lead to a misalignment between the floor-plan outline of the ceiling and floor, causing ground-truth layout height to be ill-defined.
    (c) \& (d) Our $\mathrm{LayoutWarp}$ (\cref{eq:layout3dwarp}) enhances data diversity beyond existing arts while preserving the sanity of the ground-truth layout.
    }
    \label{fig:layoutwarp}
\end{figure}

A recent finding~\cite{CubukZSL20} suggests that geometric transformations are especially helpful in improving model generalizability among various data augmentations.
Unfortunately, many commonly used perspective image transformations do not apply to 360\degree layout estimation.
For instance, when we apply image y-translation (\cref{fig:layoutwarp}'s (b)), the projected polygons of ceiling and floor boundaries get distorted and do not match in shape, while we rely on their alignment scaling factor to compute the ground-truth layout height.
This prompts us to design a principled way to perform geometric-based augmentations for 360\degree room layout.

Our core concept is simple---applying geometric transformations in 3D space rather than on 2D images.
We directly transform the ground-truth polygon and layout height and use backward warping to form the augmented view:
\begin{equation} \label{eq:layout3dwarp}
    \img' = \mathrm{LayoutWarp}\left(\img, \{\polyvert_i\}_{i=1}^{K}, h, \transform_{\polyvert}, \transform_h \right) ~,
\end{equation}
where $\transform_{\polyvert}$ transforms a polygon coordinate from source to destination and $\transform_h$ transforms layout height.
Existing 360\degree geometric augmentations---left-right flip, circular shifting, and PanoStretch~\cite{SunHSC19}---can all be realized via $\mathrm{LayoutWarp}$.
We can also produce more diverse augmentations by crafting the transformation function $\transform_{\polyvert}$ and $\transform_h$.
For instance, we can adjust camera height or randomly perturb the polygon vertices (\cref{fig:layoutwarp}'s (c) \& (d)) which is beyond what existing 360\degree data augmentations can achieve.
Please refer to the supplementary for the implementation detail of the backward warping and more visualizations.

\subsection{Pano layout studio}
\label{ssec:codebase}
Our codebase, \codebase, decomposes a layout-estimation system into different aspects---training recipes, backbones, decoders, and post-processing---each with modular design to facilitate future reuse and recombination.

\paragraph{Training recipes.}
Stochastic weight averaging~\cite{IzmailovPGVW18} is implemented, which stabilizes our training.
We also adopt RandAug~\cite{CubukZSL20}, a commonly used data augmentation for modern backbone models, and observe improved results.
We remove all geometric data augmentations from RandAug, as they are inapplicable to the 360 layout task.
Instead, we use the proposed $\mathrm{LayoutWarp}$ (\cref{ssec:newaug}) as our geometric data augmentation.

\paragraph{Backbones.}
In addition to the commonly used ResNet~\cite{HeZRS16}, we also benchmark several modern backbones for this task---HRNet~\cite{WangSCJDZLMTWLX21}, SwinTransformer~\cite{LiuLCHWZLG21}, and ConvNeXt~\cite{LiuMWFDX22}.

\paragraph{Decoders.}
In addition to our segmentation-based 2D decoder, we also implement recent regression-based 1D decoder baselines, which formulate the task as a per-column regression problem.
The per-column regressed values can then be directly projected to the floor to form a $W$-edge polygon.
We reproduce the 1D decoders from HorizonNet~\cite{SunHSC19}, HoHoNet~\cite{SunSC21}, LED2Net~\cite{WangYSCT21}, and LGTNet~\cite{JiangXXZ22} into our codebase.
We also tune these models with the new backbones and training recipes to establish a stronger baseline.
We couple the training losses with the decoders, as many losses are specific to layout representations.
Please refer to the supplementary for more details.

\paragraph{Post-processing.}
The estimated layout polygons typically contain more edges than the ground truth, which can be simplified by polygon simplification heuristics~\cite{SunHSC19,YangWPWSC19,JiangXXZ22}, which are also implemented in our codebase.


\section{Experiments}
\label{sec:results}

We conduct extensive experiments to demonstrate the advantages of the proposed \segtoreg, the effectiveness of our data augmentations, and the merit of using our \codebase codebase.


\subsection{Evaluation protocol}
We use the standard layout intersection over union (IoU) as the evaluation metric, with 2D IoU assessing the precision of the reconstructed layout floor plan and 3D IoU considering both layout floor plan and layout height accuracy.
We consider polygon simplification a decoupled task, so we mainly focus on the raw predicted geometry quality without applying post polygon simplification.
The post-processing may cause the numeral results to drift slightly, which is detailed in the supplementary.
We find the variance of run-to-run results with different random seeds could be significant, so we report the median results of four different training seeds instead.
We evaluate and compare our methods on four datasets, which are introduced in the following.

\paragraph{MatterportLayout dataset.}
Zou \etal~\cite{ZouSPCSWCH21} annotates ground-truth layout for a subset of the Matterport3D~\cite{ChangDFHNSSZZ17} dataset, comprising 1,647/190/458 labeled images for training, validation, and testing, respectively.
The captured rooms feature heavy object occlusion, and all the labeled layouts adhere to the Manhattan-world assumption.

\paragraph{Zillow Indoor dataset.}
ZInd~\cite{CruzHLKBK21} is the largest real-world dataset for 360\degree layout estimation.
We follow Jiang \etal~\cite{JiangXXZ22}'s setup to use the filtered `simple' and `raw' annotation subsets for all our experiments, and the train/valid/test split consisting of 24{,}882/3{,}081/3{,}170 images, respectively.
The rooms are mostly unfurnished but cover more diverse layout topology, including non-Manhattan layouts.

\paragraph{PanoContext and Stanford2D3D datasets.}
PanoContext~\cite{ZhangSTX14} and Stanford2D3D~\cite{ArmeniSZS17} are two small-scale datasets with only 514 and 552 images.
We follow Zou \etal~\cite{ZouSPCSWCH21}'s and Jiang \etal~\cite{JiangXXZ22}'s setup to combine all data from the other dataset when training on one of the datasets.
PanoContext is captured in the living environment, while Stanford2D3D is captured in the office rooms.
Both datasets only contain cuboid layout annotations.

\begin{table}
    \centering
    \begin{tabular}{@{} l @{\hskip 3pt} c | c @{\hskip 6pt} c @{}}
    \hline
    Backbone & Method & \# decoder params$\downarrow$ & 3DIoU(\%)$\uparrow$ \\
    \hline\hline
    \multirow{5}{*}{\makecell{ResNet-34\\(20M)}}
    & HorizonNet  & 52M & 80.48 \\
    & HoHoNet     & 30M & 80.45 \\
    & LED$^2$-Net & 52M & 80.48 \\
    & LGT-Net     & 92M & {\bf 81.55} \\
    & Ours        & {\bf 0.020M} & 81.08 \\
    \hline
    \multirow{2}{*}{\makecell{HRNet-18\\(9M)}}
    & LGT-Net     & 13M & 82.26 \\
    & Ours        & \gold{\bf 0.015M} & \gold{\bf 82.83} \\
    \hline
    \end{tabular}
    \caption{
    {\bf \codebase benchmark.}
    We summarize the quantitative comparison with the reproduced baselines on MatterportLayout~\cite{ZouSPCSWCH21} test set.
    Our all-MLP decoder is ultra-lightweight while still achieving comparable quality.
    The best result is achieved by our method with HRNet-18 as the backbone.
    }
    \label{tab:main}
    \vspace{-2em}
\end{table}


\subsection{Implementation details}
We adopt the training schedule of LGT-Net~\cite{JiangXXZ22}, where Adam~\cite{KingmaB14} optimizer with learning rate $1\mathrm{e}{-4}$ is employed.
Models are trained for 1{,}000 epochs for all datasets, except for the largest ZInd dataset, which is trained for 200 epochs.

We further employ SWA~\cite{IzmailovPGVW18} at the last 20\% of the epochs to stabilize training.
For the backbone and data augmentations, the basic setup involves ResNet-34~\cite{HeZRS16} with standard left-right flip, circular shifting, PanoStretch~\cite{SunHSC19}, and luminance jittering.
In the advanced setup, we employ HRNet-18~\cite{WangSCJDZLMTWLX21} as the backbone and replace luminance jittering with the modified RandAug~\cite{CubukZSL20} as image degradation augmentation.
We also employ random camera-height adjustment implemented by $\mathrm{LayoutWarp}$ in the advanced setup.
The random layout perturbation is not included as it only improves cross-dataset generalizability.
The same training setups are applied to all reproduced baselines.
Please refer to the supplementary for the hyperparameter details of our method.


\subsection{Codebase benchmark}
Our \codebase also reproduces many of the recent state-of-the-art methods, allowing us to have a fair evaluation with unified training schedules and configurations.
Specifically, we implement HorizonNet~\cite{SunHSC19}, HoHoNet~\cite{SunSC21}, LED$^2$-Net~\cite{WangYSCT21}, and LGT-Net~\cite{JiangXXZ22}, which are all regression-based models.
We also conduct hyperparameter tuning for these baselines, which are detailed in the supplementary.

The quantitative comparison is summarized in \Cref{tab:main}.
Despite being ultra-lightweight, our all-MLP decoder achieves better or comparable accuracy.
The lightweight MLP-only design is shown to be inferior when functioning as a regression-based 1D decoder~\cite{SunHSC19,SunSC21}.
We argue that powerful 1D decoders are necessary to capture the global scale for high-level per-column geometric property regression.
Conversely, our model is only responsible for predicting a low-level per-pixel floor plan density, which our `flattened' differentiable rendering algorithm in \segtoreg (\cref{ssec:segtoreg}) takes care of the transformation into the higher-level layout depth regression.
Essentially, our rendering algorithm acts as a similar purpose as the ``decoder" in the regression-based models, while the rendering does not have any additional parameters to learn and is already well-defined from the start.

Note that our rendering algorithm is very different from LED$^2$-Net~\cite{WangYSCT21} depth rendering.
LED$^2$-Net still employs 1D per-column regression for high-level layout geometry, with depth computed through ray-primitive intersection.
In contrast, our model predicts a 2D low-level per-pixel density, and our volume rendering entails ray marching on the estimated density field.
Our method achieves better accuracy with a thousand times fewer decoder parameters with the same backbone.

Interestingly, we observe that our method achieves superior results when paired with HRNet-18, whereas the regression-based LGT-Net performs better with ResNet-34.
Our experiments in the supplementary show that adding more backbone layers (\eg, HRNet-32 or ResNet-50) offers limited advantages.
The results suggest that different methods may prefer different types of backbones but rely less on increasing the backbone size.


\begin{table}
\centering
\begin{subtable}{\linewidth}
\centering
\begin{tabular}{@{} l @{\hskip 3pt} c | c @{\hskip 6pt} c @{}}
\hline
Method & Backbone & 2DIoU(\%)$\uparrow$ & 3DIoU(\%)$\uparrow$ \\
\hline\hline
LayoutNet v2~\cite{ZouSPCSWCH21} & ResNet-34 & 78.73 & 75.82 \\
DuLaNet v2~\cite{ZouSPCSWCH21}   & ResNet-50 & 78.82 & 75.05 \\
HorizonNet~\cite{SunHSC19}       & ResNet-50 & 81.71 & 79.11 \\
HorizonNet\ourimpl               & ResNet-34 & 82.85 & 80.48 \\
HoHoNet~\cite{SunSC21}           & ResNet-34 & 82.32 & 79.88 \\
HoHoNet\ourimpl                  & ResNet-34 & 82.71 & 80.45 \\
AtlantaNet~\cite{PintoreAG20}    & ResNet-50 & 82.09 & 80.02 \\
LED$^2$-Net~\cite{WangYSCT21}    & ResNet-50 & 82.61 & 80.14 \\
LED$^2$-Net\ourimpl              & ResNet-34 & 82.93 & 80.48 \\
LGT-Net~\cite{JiangXXZ22}        & ResNet-50 & \underline{83.52} & \underline{81.11} \\
LGT-Net\ourimpl                  & ResNet-34 & {\bf 84.05} & {\bf 81.55} \\
Ours\ourimpl                     & ResNet-34 & 83.39 & 81.08 \\
\hline
LGT-Net\ourimpl                  & HRNet-18  & 84.61 & 82.26 \\
Ours\ourimpl                     & HRNet-18  & \gold{\bf 85.27} & \gold{\bf 82.83} \\
\hline
\end{tabular}
\caption{
{\bf MatterportLayout~\cite{ZouSPCSWCH21} test set results.}
}
\label{tab:mp3d}
\vspace{.5em}
\end{subtable}
\hfill
\begin{subtable}{\linewidth}
\centering
\begin{tabular}{@{} l @{\hskip 3pt} c | c @{\hskip 6pt} c @{}}
\hline
Method & Backbone & 2DIoU(\%)$\uparrow$ & 3DIoU(\%)$\uparrow$ \\
\hline\hline
HorizonNet~\cite{SunHSC19}       & ResNet-50 & 90.44 & 88.59 \\
HorizonNet\ourimpl               & ResNet-34 & 91.37 & 89.56 \\
HoHoNet\ourimpl                  & ResNet-34 & 91.69 & 89.96 \\
LED$^2$-Net~\cite{WangYSCT21}    & ResNet-50 & 90.36 & 88.49 \\
LED$^2$-Net\ourimpl              & ResNet-34 & 91.59 & 89.78 \\
LGT-Net~\cite{JiangXXZ22}        & ResNet-50 & \underline{91.77} & \underline{89.95} \\
LGT-Net\ourimpl                  & ResNet-34 & {\bf 92.08} & {\bf 90.28} \\
\hline
LGT-Net\ourimpl                  & HRNet-18  & 92.39 & 90.61 \\
Ours\ourimpl                     & HRNet-18  & \gold{\bf 92.50} & \gold{\bf 90.73} \\
\hline
\end{tabular}
\caption{{\bf ZInd~\cite{CruzHLKBK21} test set results.}}
\label{tab:zind}
\vspace{.5em}
\end{subtable}
\hfill
\begin{subtable}{\linewidth}
\centering
\begin{tabular}{@{} l @{\hskip 3pt} c | c @{\hskip 6pt} c @{}}
\hline
Method & Backbone & \makecell{PanoC.\\3DIoU(\%)$\uparrow$} & \makecell{S2D3D\\3DIoU(\%)$\uparrow$} \\
\hline\hline
LayoutNet v2~\cite{ZouSPCSWCH21} & ResNet-34 & 85.02 & 82.66 \\
DuLaNet v2~\cite{ZouSPCSWCH21}   & ResNet-50 & 83.77 & \underline{\bf 86.60} \\
HorizonNet~\cite{SunHSC19}       & ResNet-50 & 82.63 & 82.72 \\
AtlantaNet~\cite{PintoreAG20}    & ResNet-50 & - & 83.94 \\
LGT-Net~\cite{JiangXXZ22}        & ResNet-50 & \underline{\bf 85.16} & 86.03 \\
\hline
LGT-Net\ourimpl                  & HRNet-18  & \gold{\bf 87.53} & 85.83 \\
Ours\ourimpl                     & HRNet-18  & 87.23 & \gold{\bf 87.24} \\
\hline
\end{tabular}
\caption{{\bf PanoContext~\cite{ZhangSTX14} and Stanford2D3D~\cite{ArmeniSZS17} test set results.}}
\label{tab:panos2d3d}
\end{subtable}
%
%
\caption{
{\bf Comparing our codebase results with other reports.}
The ``{\tiny \ourimpl}" indicates our \codebase reproduction, and we report the median of four training seeds.
The \underline{underline} marks the best performant method in previous reports. The {\bf bold} number is the best result of the basic or the advanced setup, while
the \gold{\bf highlighted} result is the best across the entire column.
}
\label{tab:results}
\end{table}

\subsection{Results}

We also compare our codebase results with the previous reports of other methods in \Cref{tab:results}.
All our results are the median of four training runs with different random seeds to mitigate the impact of run-to-run variance.

\paragraph{Result of the reproduced baselines.}
The entries with ``{\tiny \ourimpl}" in \Cref{tab:results} are reproduced by our \codebase.
Notably, our reproductions demonstrate consistent improvements compared to the original paper reports.
We attribute this enhancement to several implementation differences:
{\it i)} We incorporate SWA~\cite{IzmailovPGVW18} in our training.
{\it ii)} The different number of ResNet layers.
{\it iii)} The minor adjustments to the architecture and training losses according to our baseline tuning, which we describe in the supplementary.
Other experimental settings may also matter, \eg, some earlier methods~\cite{SunHSC19,SunSC21} are trained with much fewer epochs.

\paragraph{Result on complex layout.}
\Cref{tab:mp3d,tab:zind} summarizes the comparisons on datasets with complex room layout shapes.
Our method with HRNet backbone achieves the best accuracy, \ie, $+1.72$ and $+0.78$ 3D IoU improvements on MatterportLayout and ZInd datasets compared to the previous state-of-the-art report~\cite{JiangXXZ22}.

\paragraph{Result on cuboid layout.}
The comparisons on the two cuboid layout datasets are presented in \Cref{tab:panos2d3d}.
The cuboid layout is not the main focus of our study, so we only train our advanced setup with the reproduced LGT-Net and our method.
The best entries are all established by our codebase.
Our method is slightly behind the reproduced LGT-Net on PanoContext dataset, while we improve significantly on Stanford2D3D dataset.

\paragraph{Qualitative results.}
We defer the qualitative comparisons in the supplementary due to the page limit.


\subsection{Ablation study}

We show the effectiveness of the proposed \segtoreg and $\mathrm{LayoutWarp}$ via ablation studies.
The median results of four training runs are reported in all ablations as in previous experiments.

\paragraph{Formulation of the 2D layout prediction.}
We compare the results of formulating the 2D prediction as binary segmentation and floor-plan density in \Cref{tab:abla_seg2reg}.
In the case of binary segmentation, we treat the 2D prediction as a sigmoid logit map and apply the standard binary cross-entropy loss.
We borrow DuLa-Net's algorithm~\cite{YangWPWSC19} to convert the binary segmentation into a layout polygon.
The results suggest that reformulating the 2D prediction as a floor plan density field can significantly improve the results by $+0.41$ and $+0.84$ 2D IoU on MatterportLayout~\cite{ZouSPCSWCH21} and ZInd~\cite{CruzHLKBK21} valid split, respectively.

\begin{figure}
\centering
\begin{tabular}{@{}c@{\hskip 1pt}c@{\hskip 1pt}c@{}}
\includegraphics[width=.33\linewidth]{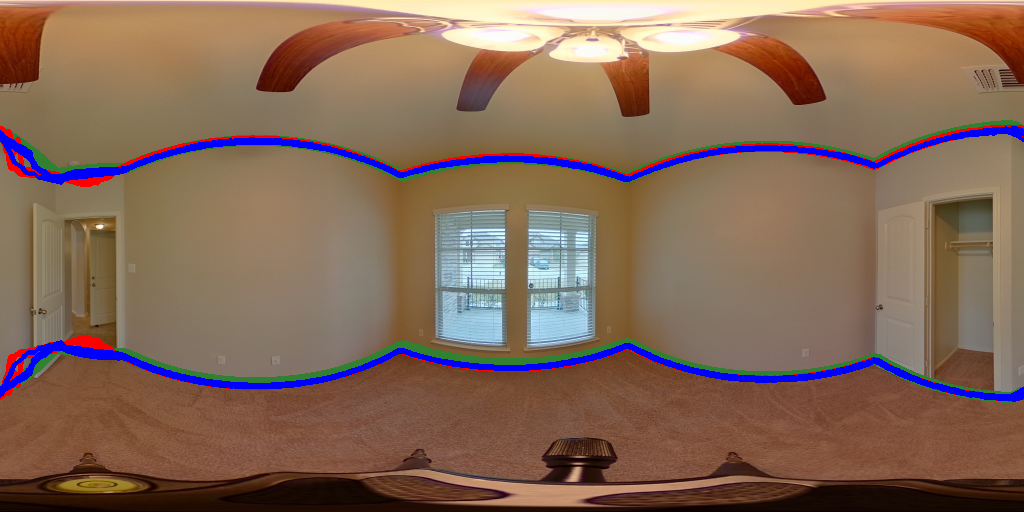} & 
\includegraphics[width=.33\linewidth]{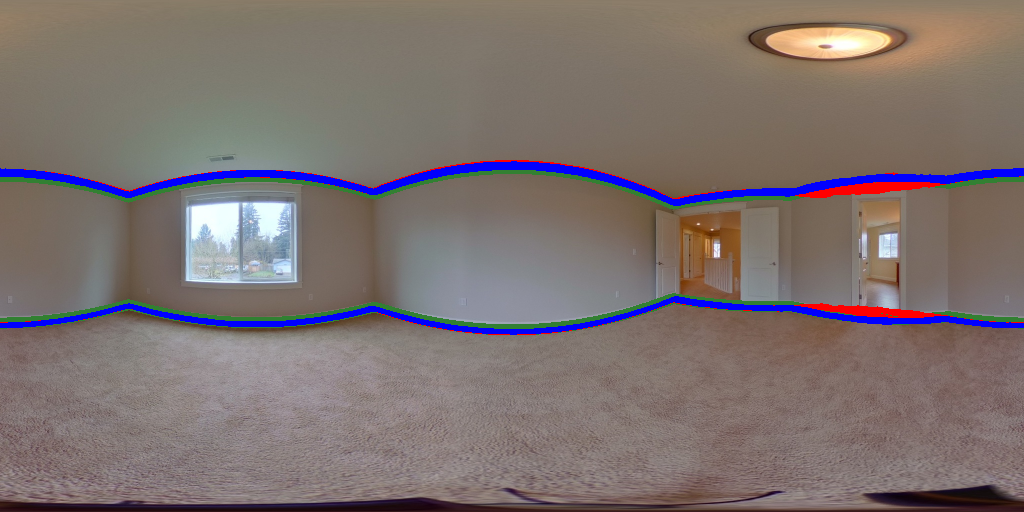} &
\includegraphics[width=.33\linewidth]{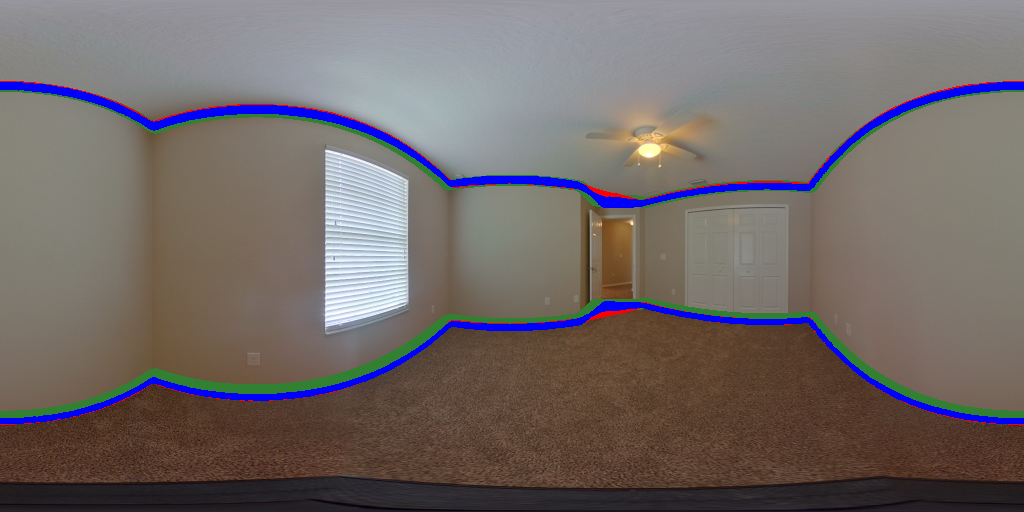} \\
\includegraphics[width=.33\linewidth]{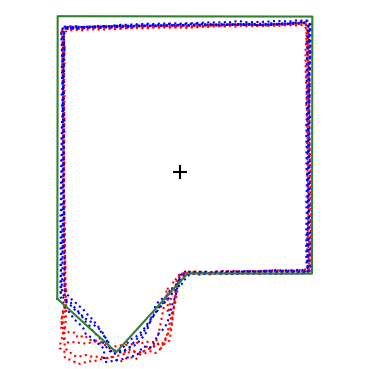} & 
\includegraphics[width=.33\linewidth]{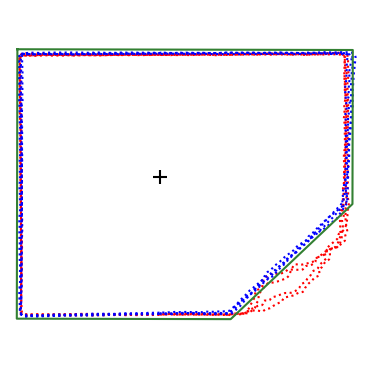} &
\includegraphics[width=.33\linewidth]{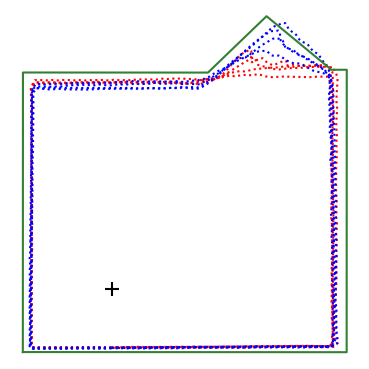} \\
\multicolumn{3}{@{}c@{}}{\textcolor[rgb]{0.01,0.75,0.24}{$\bullet$ Ground-truth} ~ \textcolor{red}{$\bullet$ Basic aug.} ~ \textcolor{blue}{$\bullet$ With rand. perturb.}}\\
\multicolumn{3}{@{}l@{}}{\scriptsize $\dagger$ The result variations are due to the four different training seeds.}
\end{tabular}
\caption{
{\bf Manhattan bias.}
The training set only consists of right-angled room layouts. The model, without random layout perturbation, `panics' when the input is not Manhattan-aligned.
}
\label{fig:manha_bias}
\vspace{-1em}
\end{figure}

\paragraph{Layout 3D warping data augmentation.}
We use the proposed $\mathrm{LayoutWarp}$ to instantiate random camera height adjustment and random layout perturbation (visualized in \cref{fig:layoutwarp}).
We train the re-produced baseline, LGT-Net, on MatterportLayout~\cite{ZouSPCSWCH21} dataset and evaluate the results on MatterportLayout and ZInd~\cite{CruzHLKBK21} valid split.
The results are shown in \Cref{tab:abla_aug}.
Random camera height adjustment achieves observable improvements on the same dataset and cross-dataset generalization.
Random layout perturbation sacrifices overall accuracy but generalizes better when non-Manhattan input is presented (the `irregular' column in \Cref{tab:abla_aug}).
The improvement may not seem apparent numerically, while the qualitative improvement is significant, as shown in \cref{fig:manha_bias}.
As MatterportLayout dataset only labels the Manhattan-aligned layout, we can clearly see that the baseline model learns the Manhattan bias, which may be helpful to infer an axis-aligned layout but is performing worse or even trying to approximate the irregular room with right-angled outlines.
In contrast, the model trained with random perturbation works more robustly in this case.
We provide more visual evidence in the supplementary.

\begin{table}
\centering
\begin{tabular}{@{} l @{\hskip 6pt} c | c @{\hskip 6pt} c @{}}
\hline
Dataset & 2D formulation & 2DIoU(\%)$\uparrow$ & 3DIoU(\%)$\uparrow$ \\
\hline\hline
\multirow{2}{*}{MpLayout}
 & binary seg. & 86.92 & 84.65 \\
 & density field & {\bf 87.33} & {\bf 85.00} \\
\hline

\multirow{2}{*}{ZInd}
 & binary seg. & 91.21 & 89.46 \\
 & density field & {\bf 92.05} & {\bf 90.39} \\
\hline
\end{tabular}
\caption{
{\bf Ablation of the \segtoreg.}
The results are reported on MatterportLayout~\cite{ZouSPCSWCH21} and ZInd~\cite{CruzHLKBK21} valid split.
The experiment shows the effectiveness of formulating the 2D per-pixel layout prediction as floor-plan density over the traditional segmentation.
}
\label{tab:abla_seg2reg}
\end{table}

\begin{table}
\centering
\begin{tabular}{@{} l | c @{\hskip 6pt} c @{\hskip 6pt} c @{}}
\hline
\multirow{3}{*}{Augmentation} & MpLayout & \multicolumn{2}{c}{MpLayout$\rightarrow$ZInd} \\
 \cline{3-4}
 & & overall & irregular \\
 & 3DIoU(\%)$\uparrow$ & 3DIoU(\%)$\uparrow$ & 3DIoU(\%)$\uparrow$ \\
\hline\hline
Basic             & 83.32 & 78.30 & 76.14 \\
w/ rand. perturb. & 82.93 & 78.77 & {\bf 77.02} \\
w/ cam. height    & {\bf 83.68} & {\bf 79.17} & 76.78 \\
\hline
\end{tabular}
\caption{
{\bf Ablation study of the new data augmentations.}
The results are measured on MatterportLayout~\cite{ZouSPCSWCH21} and ZInd~\cite{CruzHLKBK21} valid split.
We employ the reproduced LGT-Net with ResNet-34 backbone in this experiment.
See \cref{fig:layoutwarp} for visualizations of the random perturbation and camera height augmentations.
}
\label{tab:abla_aug}
\end{table}

\section{Conclusion and discussions}
\label{sec:conclusion}
We present \segtoreg, a novel approach to 360\degree room layout reconstruction that integrates the strengths of both segmentation and regression methods.
By formulating the per-pixel 2D prediction as a floor-plan density field, we can apply our `flattened' volume rendering to render layout depth regression from the density field in a differentiable and occlusion-aware manner.
Furthermore, we propose a principled method for geometric data augment on 360\degree layout task.
We also contribute \codebase, a codebase that implements several modern deep learning techniques and reproduces recent layout estimators, establishing a fair benchmark and a stronger starting point for future research.
Experimental results demonstrate that our method outperforms the previous state-of-the-art.

The success of \segtoreg highlights the potential synergy between segmentation and regression tasks in room layout estimation. 
However, we have not identified the reasons why the proposed method (or maybe all segmentation-based methods) seems to favor certain backbones.
We hope to see more studies investigating segmentation-based layout estimators in the future.
Regarding data augmentation, we instantiate two new data augmentations out of the proposed $\mathrm{LayoutWarp}$, but the potential for customization through different layout transformation functions remains largely unexplored.
We encourage future extensions to a more diverse set of 360\degree layout data augmentations.

Our method introduces volume rendering into the realm of 360\degree layout estimation, paving the way for future research to leverage and adapt techniques from the NeRF~\cite{MildenhallSTBRN20} community.
Promising avenues include exploring concepts from NeRF-based mesh reconstruction~\cite{OechsleP021,WangLLTKW21,YarivGKL21} or incorporating regularizations from few-shot NeRF~\cite{JainTA21,NiemeyerBMS0R22,YangPW23}.

{
    \small
    \bibliographystyle{ieeenat_fullname}
    \bibliography{main}

\begin{thebibliography}{55}
\providecommand{\natexlab}[1]{#1}
\providecommand{\url}[1]{\texttt{#1}}
\expandafter\ifx\csname urlstyle\endcsname\relax
  \providecommand{\doi}[1]{doi: #1}\else
  \providecommand{\doi}{doi: \begingroup \urlstyle{rm}\Url}\fi

\bibitem[Armeni et~al.(2017)Armeni, Sax, Zamir, and Savarese]{ArmeniSZS17}
Iro Armeni, Sasha Sax, Amir~R. Zamir, and Silvio Savarese.
\newblock Joint 2d-3d-semantic data for indoor scene understanding.
\newblock \emph{CoRR}, abs/1702.01105, 2017.

\bibitem[Chang et~al.(2017)Chang, Dai, Funkhouser, Halber, Nie{\ss}ner, Savva, Song, Zeng, and Zhang]{ChangDFHNSSZZ17}
Angel~X. Chang, Angela Dai, Thomas~A. Funkhouser, Maciej Halber, Matthias Nie{\ss}ner, Manolis Savva, Shuran Song, Andy Zeng, and Yinda Zhang.
\newblock Matterport3d: Learning from {RGB-D} data in indoor environments.
\newblock In \emph{2017 International Conference on 3D Vision, 3DV 2017, Qingdao, China, October 10-12, 2017}, pages 667--676. {IEEE} Computer Society, 2017.

\bibitem[Chen et~al.(2022)Chen, Xu, Geiger, Yu, and Su]{ChenXGYS22}
Anpei Chen, Zexiang Xu, Andreas Geiger, Jingyi Yu, and Hao Su.
\newblock Tensorf: Tensorial radiance fields.
\newblock In \emph{Computer Vision - {ECCV} 2022 - 17th European Conference, Tel Aviv, Israel, October 23-27, 2022, Proceedings, Part {XXXII}}, pages 333--350. Springer, 2022.

\bibitem[Cruz et~al.(2021)Cruz, Hutchcroft, Li, Khosravan, Boyadzhiev, and Kang]{CruzHLKBK21}
Steve Cruz, Will Hutchcroft, Yuguang Li, Naji Khosravan, Ivaylo Boyadzhiev, and Sing~Bing Kang.
\newblock Zillow indoor dataset: Annotated floor plans with 360deg panoramas and 3d room layouts.
\newblock In \emph{{IEEE} Conference on Computer Vision and Pattern Recognition, {CVPR} 2021, virtual, June 19-25, 2021}, pages 2133--2143. Computer Vision Foundation / {IEEE}, 2021.

\bibitem[Cubuk et~al.(2019)Cubuk, Zoph, Man{\'{e}}, Vasudevan, and Le]{CubukZMVL19}
Ekin~D. Cubuk, Barret Zoph, Dandelion Man{\'{e}}, Vijay Vasudevan, and Quoc~V. Le.
\newblock Autoaugment: Learning augmentation strategies from data.
\newblock In \emph{{IEEE} Conference on Computer Vision and Pattern Recognition, {CVPR} 2019, Long Beach, CA, USA, June 16-20, 2019}, pages 113--123. Computer Vision Foundation / {IEEE}, 2019.

\bibitem[Cubuk et~al.(2020)Cubuk, Zoph, Shlens, and Le]{CubukZSL20}
Ekin~D. Cubuk, Barret Zoph, Jonathon Shlens, and Quoc~V. Le.
\newblock Randaugment: Practical automated data augmentation with a reduced search space.
\newblock In \emph{2020 {IEEE/CVF} Conference on Computer Vision and Pattern Recognition, {CVPR} Workshops 2020, Seattle, WA, USA, June 14-19, 2020}, pages 3008--3017. Computer Vision Foundation / {IEEE}, 2020.

\bibitem[Dabouei et~al.(2021)Dabouei, Soleymani, Taherkhani, and Nasrabadi]{DaboueiSTN21}
Ali Dabouei, Sobhan Soleymani, Fariborz Taherkhani, and Nasser~M. Nasrabadi.
\newblock Supermix: Supervising the mixing data augmentation.
\newblock In \emph{{IEEE} Conference on Computer Vision and Pattern Recognition, {CVPR} 2021, virtual, June 19-25, 2021}, pages 13794--13803. Computer Vision Foundation / {IEEE}, 2021.

\bibitem[Fernandez{-}Labrador et~al.(2020)Fernandez{-}Labrador, F{\'{a}}cil, P{\'{e}}rez{-}Yus, Demonceaux, Civera, and Guerrero]{Fernandez-Labrador20}
Clara Fernandez{-}Labrador, Jos{\'{e}}~M. F{\'{a}}cil, Alejandro P{\'{e}}rez{-}Yus, C{\'{e}}dric Demonceaux, Javier Civera, and Jos{\'{e}}~Jes{\'{u}}s Guerrero.
\newblock Corners for layout: End-to-end layout recovery from 360 images.
\newblock \emph{{IEEE} Robotics Autom. Lett.}, 5\penalty0 (2):\penalty0 1255--1262, 2020.

\bibitem[Fridovich{-}Keil et~al.(2022)Fridovich{-}Keil, Yu, Tancik, Chen, Recht, and Kanazawa]{Fridovich-KeilY22}
Sara Fridovich{-}Keil, Alex Yu, Matthew Tancik, Qinhong Chen, Benjamin Recht, and Angjoo Kanazawa.
\newblock Plenoxels: Radiance fields without neural networks.
\newblock In \emph{{IEEE/CVF} Conference on Computer Vision and Pattern Recognition, {CVPR} 2022, New Orleans, LA, USA, June 18-24, 2022}, pages 5491--5500. {IEEE}, 2022.

\bibitem[Gao et~al.(2022)Gao, Chen, Su, and Chu]{GaoCSC22}
Chao{-}Chen Gao, Cheng{-}Hsiu Chen, Jheng{-}Wei Su, and Hung{-}Kuo Chu.
\newblock Layout-guided indoor panorama inpainting with plane-aware normalization.
\newblock In \emph{Computer Vision - {ACCV} 2022 - 16th Asian Conference on Computer Vision, Macao, China, December 4-8, 2022, Proceedings, Part {VI}}, pages 425--441. Springer, 2022.

\bibitem[He et~al.(2016)He, Zhang, Ren, and Sun]{HeZRS16}
Kaiming He, Xiangyu Zhang, Shaoqing Ren, and Jian Sun.
\newblock Deep residual learning for image recognition.
\newblock In \emph{2016 {IEEE} Conference on Computer Vision and Pattern Recognition, {CVPR} 2016, Las Vegas, NV, USA, June 27-30, 2016}, pages 770--778. {IEEE} Computer Society, 2016.

\bibitem[Hoffer et~al.(2020)Hoffer, Ben{-}Nun, Hubara, Giladi, Hoefler, and Soudry]{HofferBHGHS20}
Elad Hoffer, Tal Ben{-}Nun, Itay Hubara, Niv Giladi, Torsten Hoefler, and Daniel Soudry.
\newblock Augment your batch: Improving generalization through instance repetition.
\newblock In \emph{2020 {IEEE/CVF} Conference on Computer Vision and Pattern Recognition, {CVPR} 2020, Seattle, WA, USA, June 13-19, 2020}, pages 8126--8135. Computer Vision Foundation / {IEEE}, 2020.

\bibitem[Hsieh et~al.(2023)Hsieh, Sun, Dengale, and Sun]{abs-2309-09514}
Yu{-}Cheng Hsieh, Cheng Sun, Suraj Dengale, and Min Sun.
\newblock Panomixswap panorama mixing via structural swapping for indoor scene understanding.
\newblock \emph{CoRR}, abs/2309.09514, 2023.

\bibitem[Hutchcroft et~al.(2022)Hutchcroft, Li, Boyadzhiev, Wan, Wang, and Kang]{HutchcroftLBWWK22}
Will Hutchcroft, Yuguang Li, Ivaylo Boyadzhiev, Zhiqiang Wan, Haiyan Wang, and Sing~Bing Kang.
\newblock Covispose: Co-visibility pose transformer for wide-baseline relative pose estimation in 360{\textdollar}{\^{}}{\textbackslash}circ {\textdollar} indoor panoramas.
\newblock In \emph{Computer Vision - {ECCV} 2022 - 17th European Conference, Tel Aviv, Israel, October 23-27, 2022, Proceedings, Part {XXXII}}, pages 615--633. Springer, 2022.

\bibitem[Izadinia et~al.(2017)Izadinia, Shan, and Seitz]{IzadiniaSS17}
Hamid Izadinia, Qi Shan, and Steven~M. Seitz.
\newblock {IM2CAD}.
\newblock In \emph{2017 {IEEE} Conference on Computer Vision and Pattern Recognition, {CVPR} 2017, Honolulu, HI, USA, July 21-26, 2017}, pages 2422--2431. {IEEE} Computer Society, 2017.

\bibitem[Izmailov et~al.(2018)Izmailov, Podoprikhin, Garipov, Vetrov, and Wilson]{IzmailovPGVW18}
Pavel Izmailov, Dmitrii Podoprikhin, Timur Garipov, Dmitry~P. Vetrov, and Andrew~Gordon Wilson.
\newblock Averaging weights leads to wider optima and better generalization.
\newblock In \emph{Proceedings of the Thirty-Fourth Conference on Uncertainty in Artificial Intelligence, {UAI} 2018, Monterey, California, USA, August 6-10, 2018}, pages 876--885. {AUAI} Press, 2018.

\bibitem[Jain et~al.(2021)Jain, Tancik, and Abbeel]{JainTA21}
Ajay Jain, Matthew Tancik, and Pieter Abbeel.
\newblock Putting nerf on a diet: Semantically consistent few-shot view synthesis.
\newblock In \emph{2021 {IEEE/CVF} International Conference on Computer Vision, {ICCV} 2021, Montreal, QC, Canada, October 10-17, 2021}, pages 5865--5874. {IEEE}, 2021.

\bibitem[Jiang et~al.(2022)Jiang, Xiang, Xu, and Zhao]{JiangXXZ22}
Zhigang Jiang, Zhongzheng Xiang, Jinhua Xu, and Ming Zhao.
\newblock Lgt-net: Indoor panoramic room layout estimation with geometry-aware transformer network.
\newblock In \emph{{IEEE/CVF} Conference on Computer Vision and Pattern Recognition, {CVPR} 2022, New Orleans, LA, USA, June 18-24, 2022}, pages 1644--1653. {IEEE}, 2022.

\bibitem[Kajiya and Herzen(1984)]{KajiyaH84}
James~T. Kajiya and Brian~Von Herzen.
\newblock Ray tracing volume densities.
\newblock In \emph{Proceedings of the 11th Annual Conference on Computer Graphics and Interactive Techniques, {SIGGRAPH} 1984, Minneapolis, Minnesota, USA, July 23-27, 1984}, pages 165--174. {ACM}, 1984.

\bibitem[Kingma and Ba(2015)]{KingmaB14}
Diederik~P. Kingma and Jimmy Ba.
\newblock Adam: {A} method for stochastic optimization.
\newblock In \emph{3rd International Conference on Learning Representations, {ICLR} 2015, San Diego, CA, USA, May 7-9, 2015, Conference Track Proceedings}, 2015.

\bibitem[Lee et~al.(2017)Lee, Badrinarayanan, Malisiewicz, and Rabinovich]{LeeBMR17}
Chen{-}Yu Lee, Vijay Badrinarayanan, Tomasz Malisiewicz, and Andrew Rabinovich.
\newblock Roomnet: End-to-end room layout estimation.
\newblock In \emph{{IEEE} International Conference on Computer Vision, {ICCV} 2017, Venice, Italy, October 22-29, 2017}, pages 4875--4884. {IEEE} Computer Society, 2017.

\bibitem[Li et~al.(2021)Li, Yu, Sang, Wang, Song, Liu, Yeh, Zhu, Gundavarapu, Shi, Bi, Yu, Xu, Sunkavalli, Hasan, Ramamoorthi, and Chandraker]{LiYSWSLYZGSBYXS21}
Zhengqin Li, Ting{-}Wei Yu, Shen Sang, Sarah Wang, Meng Song, Yuhan Liu, Yu{-}Ying Yeh, Rui Zhu, Nitesh~B. Gundavarapu, Jia Shi, Sai Bi, Hong{-}Xing Yu, Zexiang Xu, Kalyan Sunkavalli, Milos Hasan, Ravi Ramamoorthi, and Manmohan Chandraker.
\newblock Openrooms: An open framework for photorealistic indoor scene datasets.
\newblock In \emph{{IEEE} Conference on Computer Vision and Pattern Recognition, {CVPR} 2021, virtual, June 19-25, 2021}, pages 7190--7199. Computer Vision Foundation / {IEEE}, 2021.

\bibitem[Lim et~al.(2019)Lim, Kim, Kim, Kim, and Kim]{LimKKKK19}
Sungbin Lim, Ildoo Kim, Taesup Kim, Chiheon Kim, and Sungwoong Kim.
\newblock Fast autoaugment.
\newblock In \emph{Advances in Neural Information Processing Systems 32: Annual Conference on Neural Information Processing Systems 2019, NeurIPS 2019, December 8-14, 2019, Vancouver, BC, Canada}, pages 6662--6672, 2019.

\bibitem[Liu et~al.(2021)Liu, Lin, Cao, Hu, Wei, Zhang, Lin, and Guo]{LiuLCHWZLG21}
Ze Liu, Yutong Lin, Yue Cao, Han Hu, Yixuan Wei, Zheng Zhang, Stephen Lin, and Baining Guo.
\newblock Swin transformer: Hierarchical vision transformer using shifted windows.
\newblock In \emph{2021 {IEEE/CVF} International Conference on Computer Vision, {ICCV} 2021, Montreal, QC, Canada, October 10-17, 2021}, pages 9992--10002. {IEEE}, 2021.

\bibitem[Liu et~al.(2022)Liu, Mao, Wu, Feichtenhofer, Darrell, and Xie]{LiuMWFDX22}
Zhuang Liu, Hanzi Mao, Chao{-}Yuan Wu, Christoph Feichtenhofer, Trevor Darrell, and Saining Xie.
\newblock A convnet for the 2020s.
\newblock In \emph{{IEEE/CVF} Conference on Computer Vision and Pattern Recognition, {CVPR} 2022, New Orleans, LA, USA, June 18-24, 2022}, pages 11966--11976. {IEEE}, 2022.

\bibitem[Max(1995)]{Max95a}
Nelson~L. Max.
\newblock Optical models for direct volume rendering.
\newblock \emph{{IEEE} Trans. Vis. Comput. Graph.}, 1\penalty0 (2):\penalty0 99--108, 1995.

\bibitem[Mildenhall et~al.(2020)Mildenhall, Srinivasan, Tancik, Barron, Ramamoorthi, and Ng]{MildenhallSTBRN20}
Ben Mildenhall, Pratul~P. Srinivasan, Matthew Tancik, Jonathan~T. Barron, Ravi Ramamoorthi, and Ren Ng.
\newblock Nerf: Representing scenes as neural radiance fields for view synthesis.
\newblock In \emph{Computer Vision - {ECCV} 2020 - 16th European Conference, Glasgow, UK, August 23-28, 2020, Proceedings, Part {I}}, pages 405--421. Springer, 2020.

\bibitem[M{\"{u}}ller et~al.(2022)M{\"{u}}ller, Evans, Schied, and Keller]{MullerESK22}
Thomas M{\"{u}}ller, Alex Evans, Christoph Schied, and Alexander Keller.
\newblock Instant neural graphics primitives with a multiresolution hash encoding.
\newblock \emph{{ACM} Trans. Graph.}, 41\penalty0 (4):\penalty0 102:1--102:15, 2022.

\bibitem[Niemeyer et~al.(2022)Niemeyer, Barron, Mildenhall, Sajjadi, Geiger, and Radwan]{NiemeyerBMS0R22}
Michael Niemeyer, Jonathan~T. Barron, Ben Mildenhall, Mehdi S.~M. Sajjadi, Andreas Geiger, and Noha Radwan.
\newblock Regnerf: Regularizing neural radiance fields for view synthesis from sparse inputs.
\newblock In \emph{{IEEE/CVF} Conference on Computer Vision and Pattern Recognition, {CVPR} 2022, New Orleans, LA, USA, June 18-24, 2022}, pages 5470--5480. {IEEE}, 2022.

\bibitem[Oechsle et~al.(2021)Oechsle, Peng, and Geiger]{OechsleP021}
Michael Oechsle, Songyou Peng, and Andreas Geiger.
\newblock {UNISURF:} unifying neural implicit surfaces and radiance fields for multi-view reconstruction.
\newblock In \emph{2021 {IEEE/CVF} International Conference on Computer Vision, {ICCV} 2021, Montreal, QC, Canada, October 10-17, 2021}, pages 5569--5579. {IEEE}, 2021.

\bibitem[Pintore et~al.(2020)Pintore, Agus, and Gobbetti]{PintoreAG20}
Giovanni Pintore, Marco Agus, and Enrico Gobbetti.
\newblock Atlantanet: Inferring the 3d indoor layout from a single {\textdollar}360{\^{}}{\textbackslash}circ {\textdollar} image beyond the manhattan world assumption.
\newblock In \emph{Computer Vision - {ECCV} 2020 - 16th European Conference, Glasgow, UK, August 23-28, 2020, Proceedings, Part {VIII}}, pages 432--448. Springer, 2020.

\bibitem[Schuster and Paliwal(1997)]{SchusterP97}
Mike Schuster and Kuldip~K Paliwal.
\newblock Bidirectional recurrent neural networks.
\newblock In \emph{IEEE Transactions on Signal Processing}, 1997.

\bibitem[Shabani et~al.(2021)Shabani, Song, Odamaki, Fujiki, and Furukawa]{ShabaniSOFF21}
Mohammad~Amin Shabani, Weilian Song, Makoto Odamaki, Hirochika Fujiki, and Yasutaka Furukawa.
\newblock Extreme structure from motion for indoor panoramas without visual overlaps.
\newblock In \emph{2021 {IEEE/CVF} International Conference on Computer Vision, {ICCV} 2021, Montreal, QC, Canada, October 10-17, 2021}, pages 5683--5691. {IEEE}, 2021.

\bibitem[Solarte et~al.(2022)Solarte, Liu, Wu, Tsai, and Sun]{SolarteLWTS22}
Bolivar Solarte, Yueh{-}Cheng Liu, Chin{-}Hsuan Wu, Yi{-}Hsuan Tsai, and Min Sun.
\newblock 360-dfpe: Leveraging monocular 360-layouts for direct floor plan estimation.
\newblock \emph{{IEEE} Robotics Autom. Lett.}, 7\penalty0 (3):\penalty0 6503--6510, 2022.

\bibitem[Sun et~al.(2019)Sun, Hsiao, Sun, and Chen]{SunHSC19}
Cheng Sun, Chi{-}Wei Hsiao, Min Sun, and Hwann{-}Tzong Chen.
\newblock Horizonnet: Learning room layout with 1d representation and pano stretch data augmentation.
\newblock In \emph{{IEEE} Conference on Computer Vision and Pattern Recognition, {CVPR} 2019, Long Beach, CA, USA, June 16-20, 2019}, pages 1047--1056. Computer Vision Foundation / {IEEE}, 2019.

\bibitem[Sun et~al.(2021)Sun, Sun, and Chen]{SunSC21}
Cheng Sun, Min Sun, and Hwann{-}Tzong Chen.
\newblock Hohonet: 360 indoor holistic understanding with latent horizontal features.
\newblock In \emph{{IEEE} Conference on Computer Vision and Pattern Recognition, {CVPR} 2021, virtual, June 19-25, 2021}, pages 2573--2582. Computer Vision Foundation / {IEEE}, 2021.

\bibitem[Sun et~al.(2022)Sun, Sun, and Chen]{SunSC22}
Cheng Sun, Min Sun, and Hwann{-}Tzong Chen.
\newblock Direct voxel grid optimization: Super-fast convergence for radiance fields reconstruction.
\newblock In \emph{{IEEE/CVF} Conference on Computer Vision and Pattern Recognition, {CVPR} 2022, New Orleans, LA, USA, June 18-24, 2022}, pages 5449--5459. {IEEE}, 2022.

\bibitem[Vaswani et~al.(2017)Vaswani, Shazeer, Parmar, Uszkoreit, Jones, Gomez, Kaiser, and Polosukhin]{VaswaniSPUJGKP17}
Ashish Vaswani, Noam Shazeer, Niki Parmar, Jakob Uszkoreit, Llion Jones, Aidan~N. Gomez, Lukasz Kaiser, and Illia Polosukhin.
\newblock Attention is all you need.
\newblock In \emph{Advances in Neural Information Processing Systems 30: Annual Conference on Neural Information Processing Systems 2017, 4-9 December 2017, Long Beach, CA, {USA}}, pages 5998--6008, 2017.

\bibitem[Wang et~al.(2021{\natexlab{a}})Wang, Yeh, Sun, Chiu, and Tsai]{WangYSCT21}
Fu{-}En Wang, Yu{-}Hsuan Yeh, Min Sun, Wei{-}Chen Chiu, and Yi{-}Hsuan Tsai.
\newblock Led2-net: Monocular 360deg layout estimation via differentiable depth rendering.
\newblock In \emph{{IEEE} Conference on Computer Vision and Pattern Recognition, {CVPR} 2021, virtual, June 19-25, 2021}, pages 12956--12965. Computer Vision Foundation / {IEEE}, 2021{\natexlab{a}}.

\bibitem[Wang et~al.(2021{\natexlab{b}})Wang, Sun, Cheng, Jiang, Deng, Zhao, Liu, Mu, Tan, Wang, Liu, and Xiao]{WangSCJDZLMTWLX21}
Jingdong Wang, Ke Sun, Tianheng Cheng, Borui Jiang, Chaorui Deng, Yang Zhao, Dong Liu, Yadong Mu, Mingkui Tan, Xinggang Wang, Wenyu Liu, and Bin Xiao.
\newblock Deep high-resolution representation learning for visual recognition.
\newblock \emph{{IEEE} Trans. Pattern Anal. Mach. Intell.}, 43\penalty0 (10):\penalty0 3349--3364, 2021{\natexlab{b}}.

\bibitem[Wang et~al.(2021{\natexlab{c}})Wang, Liu, Liu, Theobalt, Komura, and Wang]{WangLLTKW21}
Peng Wang, Lingjie Liu, Yuan Liu, Christian Theobalt, Taku Komura, and Wenping Wang.
\newblock Neus: Learning neural implicit surfaces by volume rendering for multi-view reconstruction.
\newblock In \emph{Advances in Neural Information Processing Systems 34: Annual Conference on Neural Information Processing Systems 2021, NeurIPS 2021, December 6-14, 2021, virtual}, pages 27171--27183, 2021{\natexlab{c}}.

\bibitem[Xie et~al.(2021)Xie, Wang, Yu, Anandkumar, {\'{A}}lvarez, and Luo]{XieWYAAL21}
Enze Xie, Wenhai Wang, Zhiding Yu, Anima Anandkumar, Jos{\'{e}}~M. {\'{A}}lvarez, and Ping Luo.
\newblock Segformer: Simple and efficient design for semantic segmentation with transformers.
\newblock In \emph{Advances in Neural Information Processing Systems 34: Annual Conference on Neural Information Processing Systems 2021, NeurIPS 2021, December 6-14, 2021, virtual}, pages 12077--12090, 2021.

\bibitem[Xu et~al.(2021)Xu, Zheng, Xu, Tang, and Gao]{XuZXTG21}
Jiale Xu, Jia Zheng, Yanyu Xu, Rui Tang, and Shenghua Gao.
\newblock Layout-guided novel view synthesis from a single indoor panorama.
\newblock In \emph{{IEEE} Conference on Computer Vision and Pattern Recognition, {CVPR} 2021, virtual, June 19-25, 2021}, pages 16438--16447. Computer Vision Foundation / {IEEE}, 2021.

\bibitem[Yang et~al.(2023)Yang, Pavone, and Wang]{YangPW23}
Jiawei Yang, Marco Pavone, and Yue Wang.
\newblock Freenerf: Improving few-shot neural rendering with free frequency regularization.
\newblock In \emph{{IEEE/CVF} Conference on Computer Vision and Pattern Recognition, {CVPR} 2023, Vancouver, BC, Canada, June 17-24, 2023}, pages 8254--8263. {IEEE}, 2023.

\bibitem[Yang et~al.(2019)Yang, Wang, Peng, Wonka, Sun, and Chu]{YangWPWSC19}
Shang{-}Ta Yang, Fu{-}En Wang, Chi{-}Han Peng, Peter Wonka, Min Sun, and Hung{-}Kuo Chu.
\newblock Dula-net: {A} dual-projection network for estimating room layouts from a single {RGB} panorama.
\newblock In \emph{{IEEE} Conference on Computer Vision and Pattern Recognition, {CVPR} 2019, Long Beach, CA, USA, June 16-20, 2019}, pages 3363--3372. Computer Vision Foundation / {IEEE}, 2019.

\bibitem[Yariv et~al.(2021)Yariv, Gu, Kasten, and Lipman]{YarivGKL21}
Lior Yariv, Jiatao Gu, Yoni Kasten, and Yaron Lipman.
\newblock Volume rendering of neural implicit surfaces.
\newblock In \emph{Advances in Neural Information Processing Systems 34: Annual Conference on Neural Information Processing Systems 2021, NeurIPS 2021, December 6-14, 2021, virtual}, pages 4805--4815, 2021.

\bibitem[Yeh et~al.(2022)Yeh, Li, Hold{-}Geoffroy, Zhu, Xu, Hasan, Sunkavalli, and Chandraker]{YehLHZXHSC22}
Yu{-}Ying Yeh, Zhengqin Li, Yannick Hold{-}Geoffroy, Rui Zhu, Zexiang Xu, Milos Hasan, Kalyan Sunkavalli, and Manmohan Chandraker.
\newblock Photoscene: Photorealistic material and lighting transfer for indoor scenes.
\newblock In \emph{{IEEE/CVF} Conference on Computer Vision and Pattern Recognition, {CVPR} 2022, New Orleans, LA, USA, June 18-24, 2022}, pages 18541--18550. {IEEE}, 2022.

\bibitem[Yun et~al.(2019)Yun, Han, Chun, Oh, Yoo, and Choe]{YunHCOYC19}
Sangdoo Yun, Dongyoon Han, Sanghyuk Chun, Seong~Joon Oh, Youngjoon Yoo, and Junsuk Choe.
\newblock Cutmix: Regularization strategy to train strong classifiers with localizable features.
\newblock In \emph{2019 {IEEE/CVF} International Conference on Computer Vision, {ICCV} 2019, Seoul, Korea (South), October 27 - November 2, 2019}, pages 6022--6031. {IEEE}, 2019.

\bibitem[Zhang et~al.(2021)Zhang, Cui, Chen, Liu, Zeng, Bao, and Zhang]{ZhangCCLZB021}
Cheng Zhang, Zhaopeng Cui, Cai Chen, Shuaicheng Liu, Bing Zeng, Hujun Bao, and Yinda Zhang.
\newblock Deeppanocontext: Panoramic 3d scene understanding with holistic scene context graph and relation-based optimization.
\newblock In \emph{2021 {IEEE/CVF} International Conference on Computer Vision, {ICCV} 2021, Montreal, QC, Canada, October 10-17, 2021}, pages 12612--12621. {IEEE}, 2021.

\bibitem[Zhang et~al.(2018)Zhang, Ciss{\'{e}}, Dauphin, and Lopez{-}Paz]{ZhangCDL18}
Hongyi Zhang, Moustapha Ciss{\'{e}}, Yann~N. Dauphin, and David Lopez{-}Paz.
\newblock mixup: Beyond empirical risk minimization.
\newblock In \emph{6th International Conference on Learning Representations, {ICLR} 2018, Vancouver, BC, Canada, April 30 - May 3, 2018, Conference Track Proceedings}. OpenReview.net, 2018.

\bibitem[Zhang et~al.(2014)Zhang, Song, Tan, and Xiao]{ZhangSTX14}
Yinda Zhang, Shuran Song, Ping Tan, and Jianxiong Xiao.
\newblock Panocontext: {A} whole-room 3d context model for panoramic scene understanding.
\newblock In \emph{Computer Vision - {ECCV} 2014 - 13th European Conference, Zurich, Switzerland, September 6-12, 2014, Proceedings, Part {VI}}, pages 668--686. Springer, 2014.

\bibitem[Zhao et~al.(2017)Zhao, Lu, Yao, Guo, Chen, and Zhang]{ZhaoLYGCZ17}
Hao Zhao, Ming Lu, Anbang Yao, Yiwen Guo, Yurong Chen, and Li Zhang.
\newblock Physics inspired optimization on semantic transfer features: An alternative method for room layout estimation.
\newblock In \emph{2017 {IEEE} Conference on Computer Vision and Pattern Recognition, {CVPR} 2017, Honolulu, HI, USA, July 21-26, 2017}, pages 870--878. {IEEE} Computer Society, 2017.

\bibitem[Zhong et~al.(2020)Zhong, Zheng, Kang, Li, and Yang]{Zhong0KL020}
Zhun Zhong, Liang Zheng, Guoliang Kang, Shaozi Li, and Yi Yang.
\newblock Random erasing data augmentation.
\newblock In \emph{The Thirty-Fourth {AAAI} Conference on Artificial Intelligence, {AAAI} 2020, The Thirty-Second Innovative Applications of Artificial Intelligence Conference, {IAAI} 2020, The Tenth {AAAI} Symposium on Educational Advances in Artificial Intelligence, {EAAI} 2020, New York, NY, USA, February 7-12, 2020}, pages 13001--13008. {AAAI} Press, 2020.

\bibitem[Zou et~al.(2018)Zou, Colburn, Shan, and Hoiem]{ZouCSH18}
Chuhang Zou, Alex Colburn, Qi Shan, and Derek Hoiem.
\newblock Layoutnet: Reconstructing the 3d room layout from a single {RGB} image.
\newblock In \emph{2018 {IEEE} Conference on Computer Vision and Pattern Recognition, {CVPR} 2018, Salt Lake City, UT, USA, June 18-22, 2018}, pages 2051--2059. Computer Vision Foundation / {IEEE} Computer Society, 2018.

\bibitem[Zou et~al.(2021)Zou, Su, Peng, Colburn, Shan, Wonka, Chu, and Hoiem]{ZouSPCSWCH21}
Chuhang Zou, Jheng{-}Wei Su, Chi{-}Han Peng, Alex Colburn, Qi Shan, Peter Wonka, Hung{-}Kuo Chu, and Derek Hoiem.
\newblock Manhattan room layout reconstruction from a single {\textdollar}360{\^{}}\{{\textbackslash}circ \}{\textdollar} image: {A} comparative study of state-of-the-art methods.
\newblock \emph{Int. J. Comput. Vis.}, 129\penalty0 (5):\penalty0 1410--1431, 2021.

\end{thebibliography}
}

\clearpage
\maketitlesupplementary


The coordinate system used in this work is explained in detail in \cref{sec:coord}.
In \cref{sec:baseline}, we show the results under various setups of the baseline models reproduced by our \codebase.
Additional technical details about layout 3D warping and \segtoreg can be found in \cref{sec:layoutwarp_detail} and \cref{sec:seg2reg_detail}.
Finally, we present an extensive qualitative comparison in \cref{sec:qual_comp}.

\section{Coordinate system}
\label{sec:coord}

As we only consider a single image, the camera position is set as the world origin.
We use a $z$-down positive world coordinate system where the positive $z$ points toward the floor.

\paragraph{Image to world.}
Let $i, j$ be the image row and column index of a pixel. We can transform them into spherical coordinates by
\begin{subequations} \label{eq:ij2uv}
\begin{align}
    u &= \left(\frac{(j + 0.5)}{W} - 0.5\right) \cdot 2\pi ~, \\
    v &= \left(\frac{(i + 0.5)}{H} - 0.5\right) \cdot \pi ~,
\end{align}
\end{subequations}
where $u$ is azimuthal angle and $v$ is the angel with respect to the $xy$-plane.
We can then lift the pixel to 3D by
\begin{subequations} \label{eq:uv2xyz}
\begin{align}
    x &= d \cdot \cos(v) \sin(u) ~, \\
    y &= d \cdot \cos(v) \cos(u) ~, \\
    z &= d \cdot \sin(v) ~,
\end{align}
\end{subequations}
where $d$ is the pixel depth.

\paragraph{World to image.}
The inverse transformation of \cref{eq:uv2xyz} is
\begin{subequations} \label{eq:xyz2uv}
\begin{align}
    u &= \mathrm{arctan2}(x, y) ~, \\
    v &= \mathrm{arctan2}\left(z, \sqrt{x^2 + y^2}\right) ~,
\end{align}
\end{subequations}
which is used in the operation $\eqproj$ (main paper Eq.~(2d)) to project sampled 3D points on the floor or the ceiling planes back to the equirectangular image for density interpolation.

\section{Baseline tuning}
\label{sec:baseline}
We implement HorizonNet~\cite{SunHSC19}, HoHoNet~\cite{SunSC21}, LED2Net~\cite{WangYSCT21}, and LGTNet~\cite{JiangXXZ22} into our codebase---\codebase.
In this section, we explore various setups of these reproduced baselines and evaluate on MatterportLayout~\cite{ZouSPCSWCH21} valid set.
We use a unified training recipe for all the methods---Adam optimizer with $1\mathrm{e}{-4}$ learning rate trained for 1k epochs.
We activate Stochastic Weight Averaging~\cite{IzmailovPGVW18} in the last 200 epochs to stabilize training.
Except stated otherwise, we use the basic setup with LGT-Net~\cite{JiangXXZ22}---ResNet-34 as backbone; standard left-right flip, circular shifting, PanoStretch~\cite{SunHSC19}, and luminance jittering as data augmentation.
We accumulate the results from different training seeds as box plots.

\begin{figure}
    \centering
    \includegraphics[width=.8\linewidth]{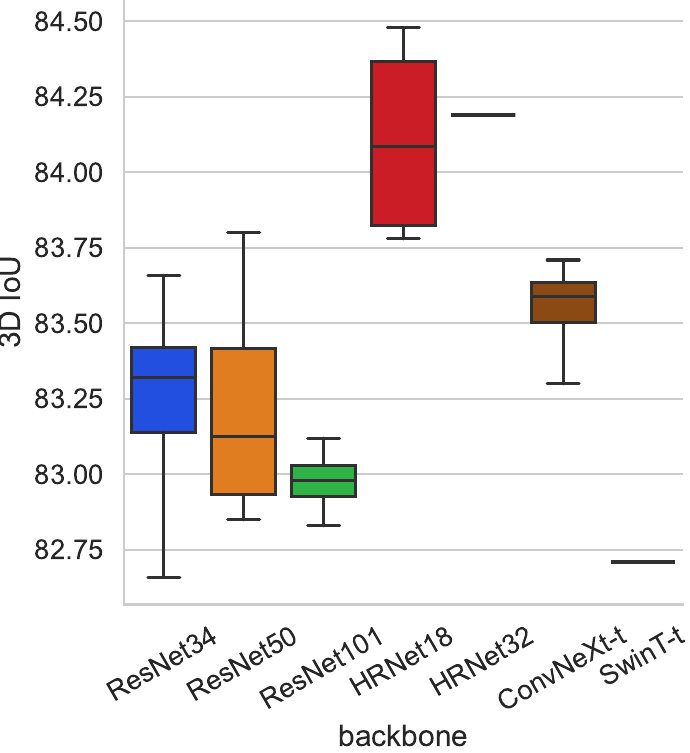}
    \caption{
    {\bf Results of different backbones.}
    The results of HRNet32 and SwinT-t are obtained from only one training seed.
    }
    \label{fig:baselines_backbone}
\end{figure}

\paragraph{Backbones.}
We show the results with different backbones, including ResNet~\cite{HeZRS16} and the more advanced HRNet~\cite{WangSCJDZLMTWLX21}, ConvNeXt~\cite{LiuMWFDX22}, and SwinTransformer~\cite{LiuLCHWZLG21}, in \cref{fig:baselines_backbone}.
Overall, HRNet performs especially well in this task; ConvNeXt is slightly above ResNet; the SwinTransformer backbone seems to be unsuitable for this task.
Increasing the number of backbone layers offers limited merits.
The result with ResNet101 is even worse than ResNet34.
Please note that all the results are trained with the same recipe.
It could be possible that larger models or different network architectures need different training recipes.
More future research about the different network architectures for this task would be valuable.

\begin{figure*}
    \centering
    \includegraphics[width=.95\linewidth]{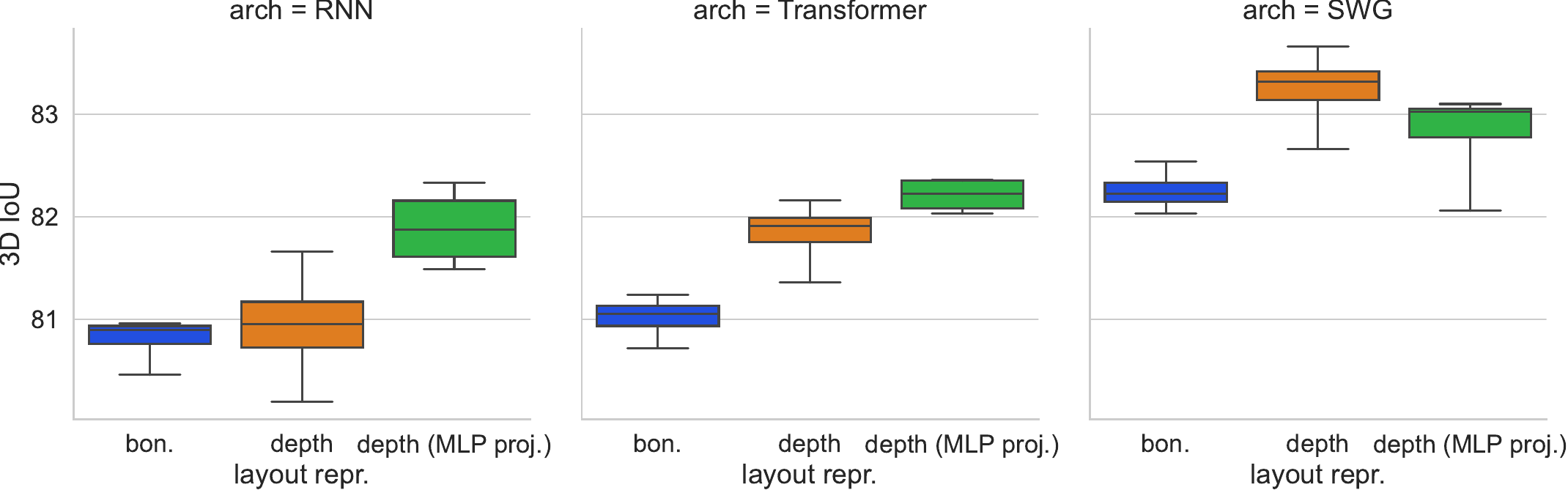}
    \caption{
    {\bf Results of the regression-based methods.}
    Originally, RNN is employed by HorizonNet, Transformer is employed by HoHoNet, and SWG is proposed by LGT-Net as the 1D decoder.
    The layout representation `bon' indicates predicting per-column layout boundary on the image space while `depth' is for per-column layout depth.
    `MLP proj.` indicates two non-linear layers are added at the very end of the decoder before predicting the layout.
    }
    \label{fig:baselines_decoder}
\end{figure*}

\begin{figure}
    \centering
    \includegraphics[width=.6\linewidth]{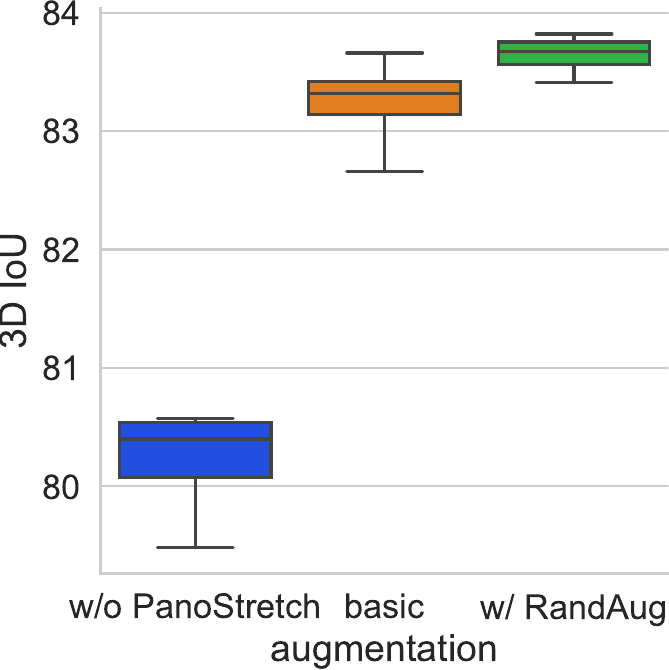}
    \caption{
    {\bf Results of modified basic data augmentations.}
    PanoStretch is crucial to recent methods to achieve state-of-the-art results.
    Employing RandAug can further improve the result.
    }
    \label{fig:baselines_aug}
\end{figure}

\paragraph{1D decoder and model head.}
We decouple the regression-based decoder into the architecture and the layout representations.
The results are presented in \cref{fig:baselines_decoder}.
The 1D network architecture of HorizonNet~\cite{SunHSC19}, HoHoNet~\cite{SunSC21}, and LGTNet~\cite{JiangXXZ22} are RNN~\cite{SchusterP97}, Transformer~\cite{VaswaniSPUJGKP17}, and SWG (an adaptation of Swin-transformer~\cite{LiuLCHWZLG21}), respectively.
HorizonNet proposes to predict per-column layout boundary on image space, while LGT-Net proposes to predict per-column layout depth with a layout height.
For the architecture, SWG significantly outperforms Transformer and RNN.
Directly predicting layout depth consistently improves the results from all architectures compared to predicting layout boundary on the image space.
Originally, the features from all the architecture are just followed by a linear projection layer to predict layout.
We also try to add two additional non-linear MLP layers, which improve RNN and Transformer decoder but degrade SWG results.

Interestingly, we find directly training SWG decoder to predict layout boundary fails to converge.
We find it is because {\it i)} the output variance of SWG is high, and {\it ii)} the layout boundary prediction uses sigmoid to constraint the output range.
The combination of these two causes the last layer to `die' as predicting a large absolute value receives zero gradient from the sigmoid function.
As a workaround, we multiply the output value of SWG by $0.1$ for the layout boundary prediction.
The result suggests that predicting layout boundary may be more unstable compared to predicting layout depth, which may be one of the reason for the superiority of predicting layout depth.

\begin{figure}
    \centering
    \includegraphics[width=.7\linewidth]{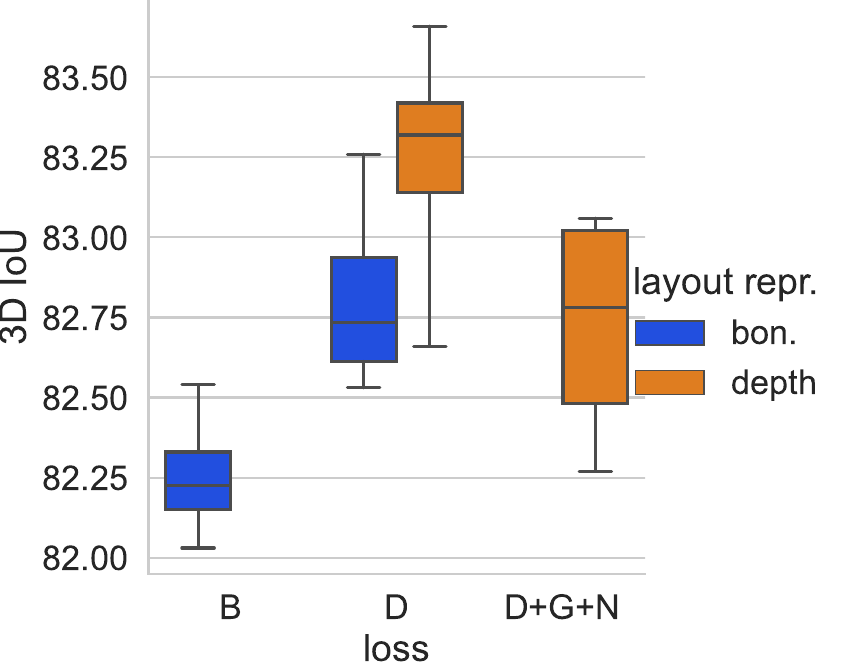}
    \caption{
    {\bf Results of regression training loss.}
    Training with layout depth loss improves results even when the model predicts layout boundary on image space.
    Directly predicting layout depth achieves the best result.
    Adding gradient and normal losses to regularize layout depth does not improve.
    }
    \label{fig:baselines_loss}
\end{figure}

\paragraph{Augmentations.}
The basic augmentation consists of left-right flip, circular shifting, PanoStretch~\cite{SunHSC19}, and luminance jittering.
In \cref{fig:baselines_aug}, we show the results of ablating PanoStretch and replacing luminance jittering with the modified modified RandAug~\cite{CubukZSL20}.
The results show that PanoStretch is crucial to the quality, highlighting the importance of geometric-based data augmentation.
Using RandAug introduces a more diverse image appearance and improves results further.

\paragraph{Losses.}
We explore existing training losses for regression-based methods.
The result is presented in \cref{fig:baselines_loss}.
LED$^2$-Net~\cite{WangYSCT21} proposes to render layout depth from the predicted layout boundary on the image space, which we find indeed can improve boundary prediction quality.
We try the gradient and normal losses employed by LGT-Net~\cite{JiangXXZ22} but do not observe improvement in our reproduction.

\begin{figure*}
    \centering
    \includegraphics[trim=0 150 0 0, clip,width=\linewidth]{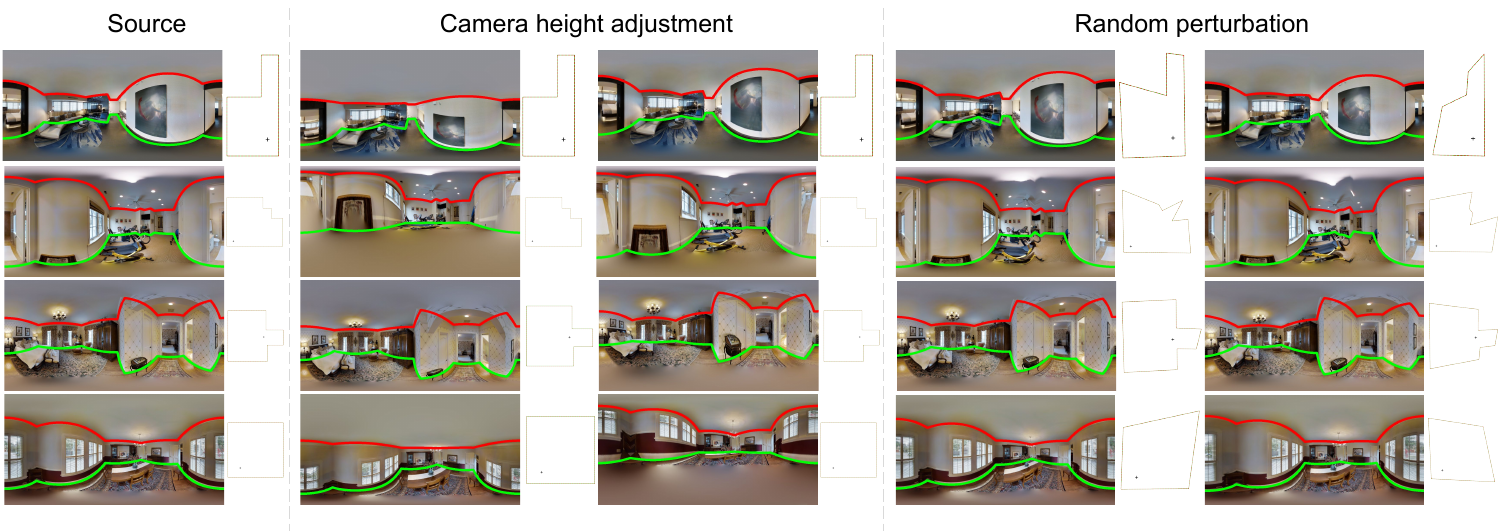}
    \caption{
        {\bf More visualization of the new data augmentations by $\mathrm{LayoutWarp}$}
    }
    \label{fig:layoutwarp_more}
\end{figure*}

\section{Layout 3D warping---more details}
\label{sec:layoutwarp_detail}
The proposed $\mathrm{LayoutWarp}$ (main paper Eq.~(16)) enables us to produce more diverse geometrically augmented views by crafting layout polygon transformation functions $\transform_{\polyvert}$ and layout height transformation functions $\transform_h$.
In the following, we detail how we compute the source image coordinates from the destination image based on the given transformation so that we can apply backward warping to form the transformed image.

\paragraph{Source image coordinate computation.}
Recap that $\mathrm{LayoutWarp}$ takes a source image $\img$ and it's layout polygon $\{\polyvert_i\}_{i=1}^{K}$, layout height $h$, and their transformation functions $\transform_{\polyvert}, \transform_h$ as input to form the warped image $\img'$:
\[
    \img' = \mathrm{LayoutWarp}\left(\img, \{\polyvert_i\}_{i=1}^{K}, h, \transform_{\polyvert}, \transform_h \right) ~.
\]
Let 
\begin{subequations} \label{eq:transpoly}
\begin{align}
    \{\polyvert_i'\}_{i=1}^{K} &= \transform_{\polyvert}\left( \{\polyvert_i\}_{i=1}^{K} \right) ~, \\
    h' &= \transform_h\left( h \right) ~,
\end{align}
\end{subequations}
where $\{\polyvert'\}_{i=1}^K$ and $h'$ are the transformed layout polygon coordinates and layout height.
To do backward warping, for each target pixel $(i', j')$ at the destination image, we want to compute its corresponding coordinate $(i, j)$ on the source image following the given layout transformation.
We first use \cref{eq:ij2uv} to compute the spherical coordinate $(u', v')$ of the target pixel.
By casting a 2D ray following the azimuthal angle $u'$, we can find the ray-polygon intersection point $\hat{\polyvert}$ on the $k$-th layout polygon edge $\overline{\polyvert'_k \polyvert'_{k+1}}$ (let $\polyvert'_{K+1}=\polyvert'_1$ as the polygon is closed).
The depth term in \cref{eq:uv2xyz} is
\begin{equation} \label{eq:depth2layout}
    d' = \begin{cases}
      \zfloor \cdot \csc(v'), & \text{if}\ (i',j')\in\text{floor} \\
      \left(\zfloor-h'\right) \cdot \csc(v'), & \text{if}\ (i',j')\in\text{ceiling} \\
      \|\hat{\polyvert}\| \cdot \sec(v'), & \text{if}\ (i',j')\in\text{wall} ~,
    \end{cases}
\end{equation}
which lifts the pixel to a 3D point $(x', y', z')$.
The corresponding source 3D point is
\begin{subequations} \label{eq:xyz_dst2src}
\begin{align}
    x &= a x' + b y' ~, \\
    y &= c x' + d y' ~, \\
    z &= \zfloor - \frac{h}{h'} \cdot (\zfloor - z') ~,
\end{align}
\end{subequations}
where the backward transformation parameters $a, b, c, d$ align $\overline{\polyvert'_k \polyvert'_{k+1}}$ to $\overline{\polyvert_k \polyvert_{k+1}}$.
Let $\polyvert'_k = (x'_k, y'_k)$ and $\polyvert_k = (x_k, y_k)$, we solve the following linear system:
\begin{equation} \label{eq:linearsys}
\begin{bmatrix}
x'_k & y'_k & 0 & 0 \\ 
0 & 0 & x'_k & y'_k \\ 
x'_{k+1} & y'_{k+1} & 0 & 0 \\ 
 0 & 0 & x'_{k+1} & y'_{k+1} \\ 
\end{bmatrix}
\begin{bmatrix}
a \\ 
b \\ 
c \\ 
d
\end{bmatrix}
=
\begin{bmatrix}
x_k \\ 
y_k \\ 
x_{k+1}\\ 
y_{k+1}
\end{bmatrix} ~,
\end{equation}
where the solution is
\begin{subequations} \label{eq:abcd}
\begin{align}
    a &= \frac{y'_{k+1} \cdot x_k - y'_k \cdot x_{k+1}}{y'_{k+1} \cdot x'_k - y'_k \cdot x'_{k+1}} ~, \\
    b &= \frac{x'_{k+1} \cdot x_k - x'_k \cdot x_{k+1}}{x'_{k+1} \cdot y'_k - x'_k \cdot y'_{k+1}} ~, \\
    c &= \frac{y'_{k+1} \cdot y_k - y'_k \cdot y_{k+1}}{y'_{k+1} \cdot x'_k - y'_k \cdot x'_{k+1}} ~, \\
    d &= \frac{x'_{k+1} \cdot y_k - x'_k \cdot y_{k+1}}{x'_{k+1} \cdot y'_k - x'_k \cdot y'_{k+1}} ~.
\end{align}
\end{subequations}
Note that we assume $\polyvert_{k}\neq\polyvert_{k+1}$ and $\polyvert'_{k}\neq\polyvert'_{k+1}$ for all layout polygon edges.
Finally, the corresponding source 3D point $(x, y, z)$ from \cref{eq:xyz_dst2src} is projected to the source image using \cref{eq:xyz2uv} to interpolate the color.

\begin{table}
    \centering
    \begin{tabular}{@{}lcc@{}}
    \hline
    Effect & $\transform_{\polyvert}$ & $\transform_h$ \\
    \midrule
    \makecell[l]{Left-right\\circular shifting} & 
        $\polyvert'_k = \begin{bmatrix}
        \cos(\theta) & -\sin(\theta) \\ 
        \sin(\theta) & \cos(\theta) 
        \end{bmatrix} \polyvert_k$ &
        $h' = h$ \\
    \multicolumn{3}{@{}l@{}}{\footnotesize $^{\dagger}$ The rotation $\theta$ is shared by all the $K$ vertices.} \\
    \midrule
    \makecell[l]{Left-right flip} & 
        $\polyvert'_k = \begin{bmatrix}
        -1 & 0 \\ 
        0 & 1
        \end{bmatrix} \polyvert_k$ &
        $h' = h$ \\
    \midrule
    \makecell[l]{PanoStretch~\cite{SunHSC19}} & 
        $\polyvert'_k = \begin{bmatrix}
        s_x & 0 \\ 
        0 & s_y
        \end{bmatrix} \polyvert_k$ &
        $h' = h$ \\
    \multicolumn{3}{@{}l@{}}{\footnotesize $^{\dagger}$ The scaling factors $s_x, s_y$ are shared by all the $K$ vertices.} \\
    \midrule
    \makecell[l]{Camera height\\adjustment} & 
        $\polyvert'_k = s \polyvert_k$ &
        $h' = s h$ \\
    \multicolumn{3}{@{}l@{}}{\footnotesize $^{\dagger}$ The $s$ scales camera height by $\frac{1}{s}$ and is shared by all vertices and height.} \\
    \midrule
    \makecell[l]{Random\\perturbation} & 
        $\polyvert'_k = s_k \polyvert_k$ &
        $h' = h$ \\
    \multicolumn{3}{@{}l@{}}{\footnotesize $^{\dagger}$ Each of the $K$ vertex has it's own scaling factor.} \\
    \hline
    \end{tabular}
    \caption{{\bf Geometric-based data augmentation by $\mathrm{LayoutWarp}$.}}
    \label{tab:layoutwarp_op}
\end{table}

\begin{figure*}
    \centering
    \begin{tabular}{@{}c@{}}
    \includegraphics[trim=0 140 0 0, clip,width=\linewidth]{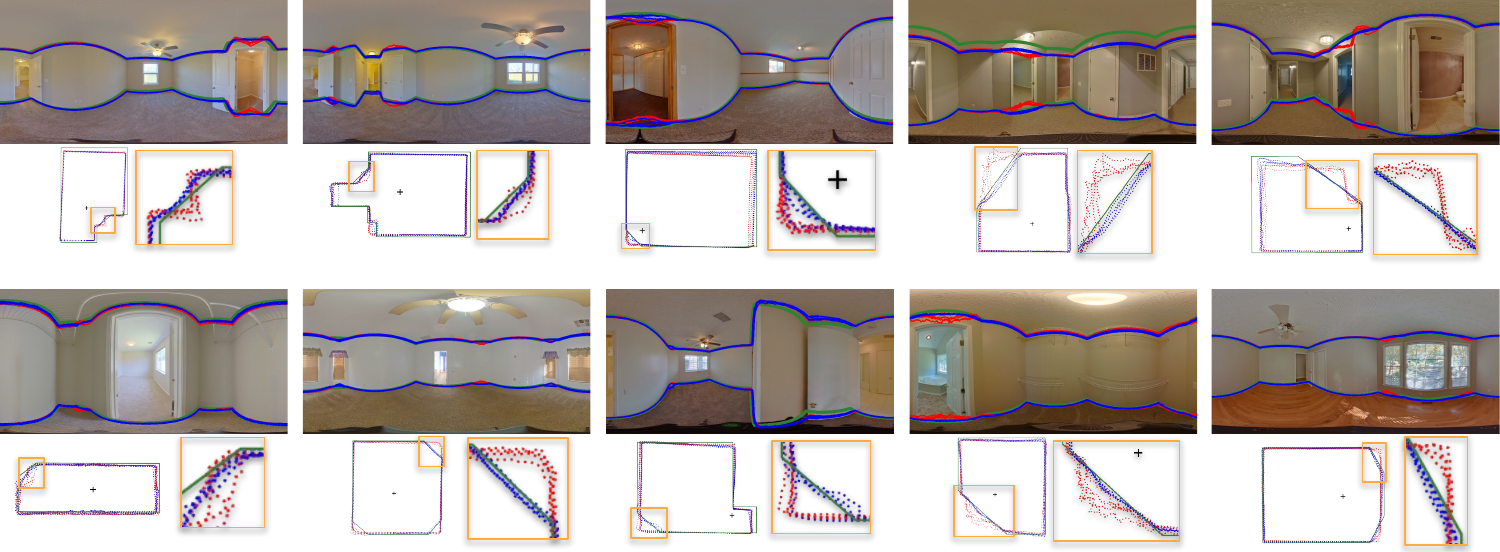}\\
    {\textcolor[rgb]{0.01,0.75,0.24}{$\bullet$ Ground-truth} ~ \textcolor{red}{$\bullet$ Basic aug.} ~ \textcolor{blue}{$\bullet$ With rand. perturb.}}\\
    {\scriptsize $\dagger$ The result variations are due to the four different training seeds.}
    \end{tabular}
    \caption{
        {\bf Training with random perturbation generalizes better from Manhattan to general layout.}
        We can observe that the models trained with random perturbation are less ‘panics’ when the input is not Manhattan-aligned.
    }
    \label{fig:manha_bias_supp}
\end{figure*}

\paragraph{Crafting the transformation functions.}
We can craft $\transform_{\polyvert}, \transform_h$ to produce various geometric augmentations.
We summarize the realization of the commonly-used left-right circular shifting, left-right flip, PanoStretch~\cite{SunHSC19}, and the new camera height adjustment and random layout polygon perturbation in \cref{tab:layoutwarp_op}.
We visualize more augmented views in \cref{fig:layoutwarp_more}.

\paragraph{More qualitative comparison of random perturbation.}
In the main paper, we show random perturbation can improve the results when generalizing from a Manhattan-aligned dataset to a general layout dataset.
We show more visual comparison in \cref{fig:manha_bias_supp}.

\section{Seg2Reg---more details}
\label{sec:seg2reg_detail}
We provide more technical details and analyses of the proposed \segtoreg in this section.

\paragraph{Training details.}
Our \segtoreg enables differentiable layout depth (\ie, distance to layout wall on the floor plan) rendering from the 2D density map prediction.
The ‘flattened’ volume rendering is detailed in the main paper Sec.~3.1.
Directly supervising the rendering by ground-truth layout depth leads to ambiguity.
Recap that the depth rendering equation of a ray is (main paper Eq.(3)):
\[
    d = \sum\nolimits_{i=1}^\npts T_i \alpha_i t_i ~, \text{where}~ T_i = \prod\nolimits_{j=1}^{i-1} \left(1 - \alpha_j\right) ~,
\]
where $K$ is the number of sampled points on the ray.
We can see that there are an infinite number of weight distributions on the ray that can render to a specific ground-truth depth $d^*$.
To resolve ambiguity, we directly derive a compact weight distribution $w^*$ for the ground-truth depth $d^*$.
Let $t_k, t_{k+1}$ be the two depth values nearest to $d^*$ on the ray.
Our ground truth weight distribution is:
\begin{subequations} \label{eq:gt_w}
\begin{align}
    w_k &= \frac{t_{k+1} - d^*}{t_{k+1} - t_k} ~, \\
    w_{k+1} &= 1 - w_k ~, \\
    w_i &= 0~, \text{where}~ i=1,\cdots,k{-}1,k{+}2,\cdots,K+1 ~.
\end{align}
\end{subequations}
There is a special case when the far clipping distance of the ray does not reach the layout $t_{K} < d^*$.
In this special case, the weights are all zero except $W_{K+1}=1$.

To render the primary layout, we directly use the $\frac{H}{2}$ points in the image column as the sampled points.
For the secondary layouts, we sample $\nsecondary=32$ secondary camera centers.
To render layout depth for the secondary cameras, we uniformly sampled $1{,}024$ points on a ray with the farthest distance $t_{K}=16$ (the camera-to-floor distance is $\zfloor=1.6$).
The loss weights for $\Lprimary$, $\Lsecondary$, and $\Lseg$ are $w_1=1.0, w_2=0.1, w_3=1.0$, respectively.

\setlength{\columnsep}{6pt}
\begin{wrapfigure}{r}{0.15\linewidth}
\centering
\vspace{-1.5em}
\includegraphics[width=1\linewidth]{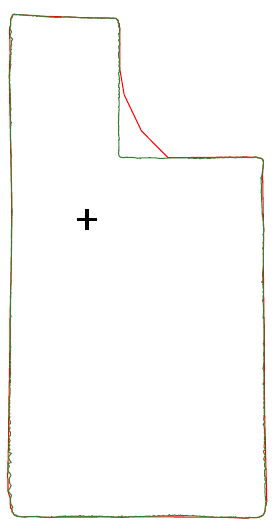}
\vspace{-2em}
\end{wrapfigure}
\paragraph{Robust polygons merging algorithm.}
To merge the primary layout polygon and secondary layout polygons, the simplest way is to directly take the polygon union.
Such a naive strategy may have polygon edges crossing the region outside the layout where the model actually predicts high floor plan density (the \textcolor{red}{red polygon} in the figure), especially when the rendered secondary polygons are in lower resolution (fewer number of vertices).
This prompts us to design a more robust algorithm.
Our idea is to connect all the rendered vertices with the minimum perimeter.
We first compute the pair-wise distance matrix of all the rendered vertices from primary and secondary views.
The pair-wise distance matrix represents a complete graph.
We then construct a minimum spanning tree from the complete graph via the Kruskal algorithm.
The final merged polygon (the \textcolor{green}{green polygon} in the figure) is determined by the trajectory connecting the farthest two points on the tree, which can be realized by running the depth-first tree search two times.

We show the results of the two polygon merging methods and the result after polygon simplification in \cref{tab:polypost}.

\begin{table}[!ht]
    \centering
    \begin{tabular}{@{}ccc@{}}
    \hline
    Union & MST & MST + simplification \\
    \hline
    87.22\% & 87.33\% & 87.13\% \\
    \hline
    \end{tabular}
    \caption{
    {\bf Results of different polygon merging algorithms.}
    We report the 2DIoU$\uparrow$ comparison on MatterportLayout valid set.
    Polygon merging via minimum spanning tree (MST) achieves slightly better results.
    Applying polygon simplification causes the number result to drop slightly.
    }
    \label{tab:polypost}
\end{table}

\begin{figure}
    \centering
    \includegraphics[trim=0 200 360 0, clip,width=\linewidth]{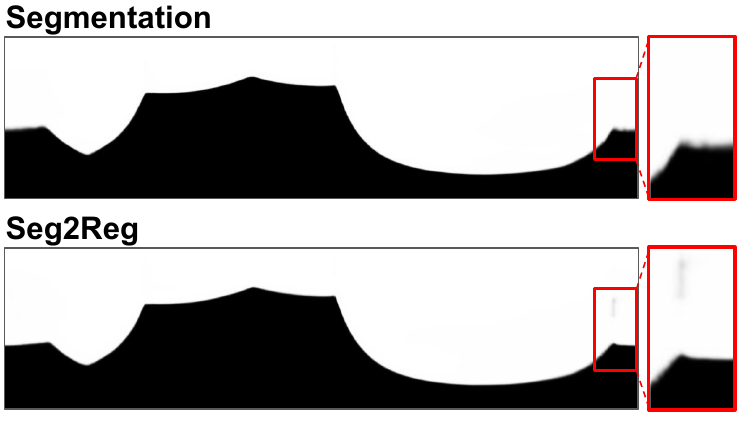}
    \caption{
    {\bf Visualization of the raw predictions.}
    Segmentation-only method and our \segtoreg converge to different results.
    }
    \label{fig:raw_pred}
\end{figure}

\paragraph{Raw prediction visualizations.}
In the main paper's Table.3, we show that our \segtoreg achieves better quantitative results than a purely segmentation-based method.
In \cref{fig:raw_pred}, we visualize the raw predictions activated by $\sigmoid$.
Segmentation-based method only supervises per-pixel classification.
The proposed \segtoreg introduces rendering loss, which emphasizes the rendered polygon position, while the density far outside the layout polygon is less important as they do not affect the rendering results.

\section{Qualitative comparison}
\label{sec:qual_comp}
\begin{figure*}
    \centering
    \begin{tabular}{@{}c@{}}
    \includegraphics[trim=0 230 0 0, clip,width=\linewidth]{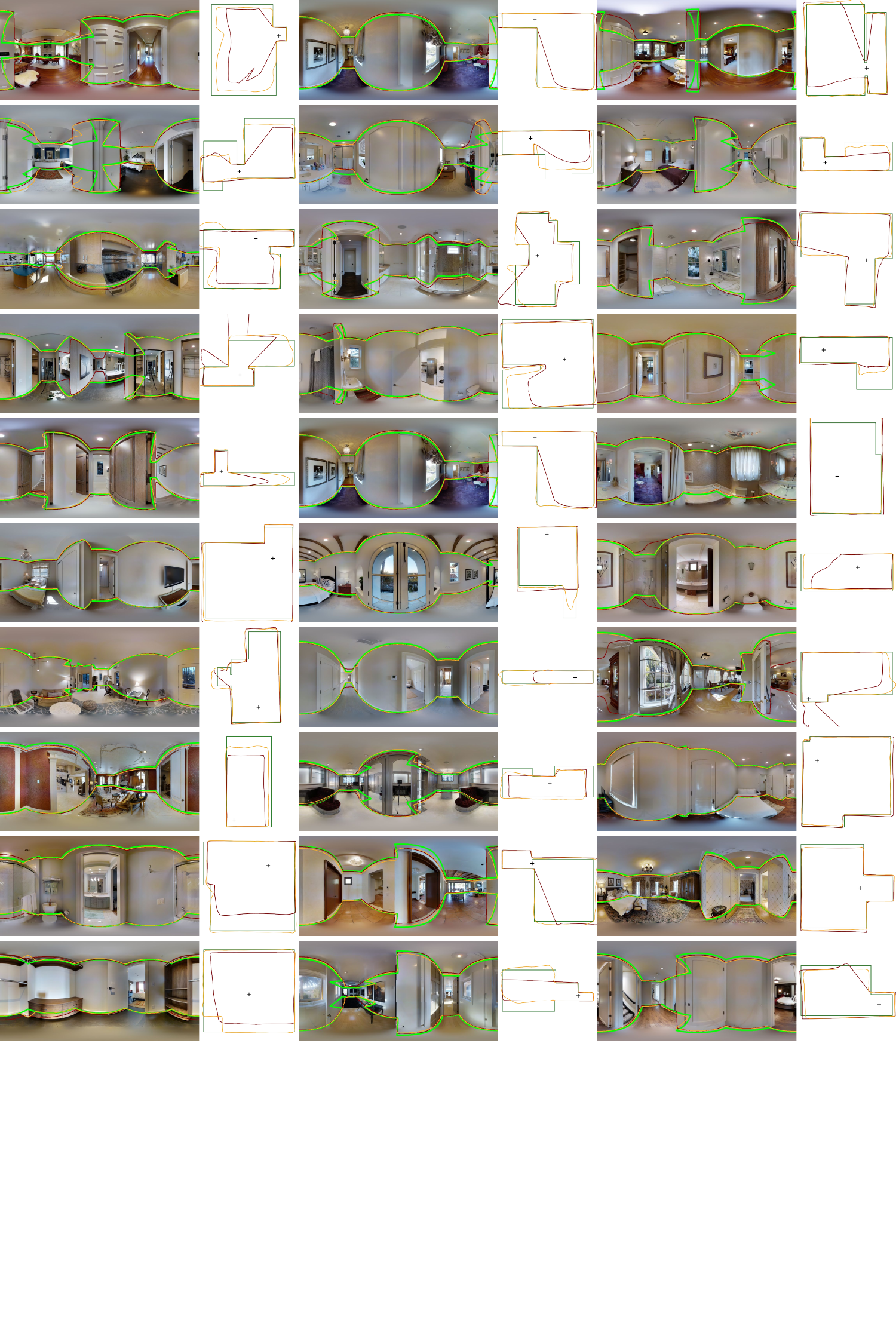}\\
    {\textcolor[rgb]{0.2, 0.5, 0.2}{$\bullet$ Ground-truth} ~ \textcolor[rgb]{1, 0.674, 0.109}{$\bullet$ \segtoreg} ~ \textcolor[rgb]{0.54, 0, 0}{$\bullet$ LGT-Net}}
    \end{tabular}
    \caption{
    {\bf Qualitative comparisons on the unseen data from MatterportLayout.}
    We compare it with the state-of-the-art model from LGT-Net's official repo.
    The test set 2DIoU is $83.52$\% for the official LGT-Net and $85.14$\% for our model.
    }
    \label{fig:qual_comp}
\end{figure*}

We provide an extensive qualitative comparison in \cref{fig:qual_comp}.
We run LGT-Net's official repo to show the performance of the existing state-of-the-art.

\end{document}